\pdfoutput=1

\documentclass[11pt]{article}

\usepackage[preprint]{acl}

\usepackage{times}
\usepackage{latexsym}

\usepackage[T1]{fontenc}

\usepackage[utf8]{inputenc}

\usepackage{microtype}

\usepackage{inconsolata}

\usepackage{graphicx}

\usepackage{algorithm}
\usepackage{algpseudocode}
\usepackage{enumitem}
\usepackage{makecell}
\usepackage{hyperref}

\newcommand{\algorithmicbreak}{\textbf{break}}
\newcommand{\Break}{\State \algorithmicbreak}

\title{{Enhancing Logical Reasoning in Large Language Models through Graph-based 
Synthetic Data}}

\author{
 Jiaming Zhou$^{1\spadesuit}$\hspace{2mm}
 \hspace{2mm}Abbas Ghaddar$^1$\hspace{2mm}Ge Zhang$^1$\hspace{2mm}Liheng Ma$^2$\hspace{2mm}Yaochen Hu$^1$\hspace{2mm} \\
 \textbf{Soumyasundar Pal$^1$\hspace{2mm}
 Mark Coates$^2$\hspace{2mm}Bin Wang$^3$\hspace{2mm}Yingxue Zhang$^{1\spadesuit}$\hspace{2mm}Jianye Hao$^3$\hspace{2mm}}
\\
$^1$ Huawei Noah’s Ark Lab, Montréal, Canada \\
$^2$ McGill University and Mila - Québec AI Institute\\
$^3$ Huawei Noah’s Ark Lab, Beijing, China \\
\texttt{\href{mailto:email@domain}{jiaming.zhou@h-partners.com},}\hspace{3mm}\texttt{\href{mailto:email@domain}{yingxue.zhang@huawei.com}}
}

\usepackage{booktabs}
\usepackage{eqparbox}
\usepackage{arydshln}
\usepackage{subcaption}
\usepackage{rotating,csquotes}
\usepackage{xcolor}
\usepackage{multirow}
\usepackage{framed}
\usepackage{booktabs}
\usepackage{amsmath}
\usepackage{amssymb}
\usepackage{mathtools}
\usepackage{upgreek}
\usepackage{soul}
\usepackage{amsfonts}
\usepackage{todonotes}
\usepackage[most]{tcolorbox}
\usepackage{tabularx}

\newtcbox{\mybox}[1][]{enhanced, colframe=blue, colback=blue!15, 
	frame style={opacity=0.25}, interior style={opacity=0.25}, 
	nobeforeafter, tcbox raise base, shrink tight, extrude by=1mm, #1}

\usepackage{amssymb}
\usepackage{pifont}

\DeclareMathAlphabet\mathbfcal{OMS}{cmsy}{b}{n}

\newcommand{\mightmention}[1]{}
\newcommand{\problem}[1]{\textcolor{red}{$\star$}}
\newcommand{\answer}[1]{\textcolor{blue}{$\#$}}
\newcommand{\todoreview}[1]{\textcolor{green}{$@$}}

\usepackage[most]{tcolorbox}
\newcommand{\specialcell}[2][c]{%
  \begin{tabular}[#1]{@{}c@{}}#2\end{tabular}}

\usepackage{amsmath,amsfonts,bm}

\newcommand{\qed}{\nobreak \ifvmode \relax \else
      \ifdim\lastskip<1.5em \hskip-\lastskip
      \hskip1.5em plus0em minus0.5em \fi \nobreak
      \vrule height0.75em width0.5em depth0.25em\fi}

\def\eqref#1{equation~\ref{#1}}

\def\1{\bm{1}}

\DeclareMathAlphabet{\mathsfit}{\encodingdefault}{\sfdefault}{m}{sl}
\SetMathAlphabet{\mathsfit}{bold}{\encodingdefault}{\sfdefault}{bx}{n}

\def\gG{{\mathcal{G}}}

\def\gR{{\mathcal{R}}}

\def\gV{{\mathcal{V}}}

\newcommand{\cR}{\mathcal{R}}

\DeclareMathAlphabet\mathbfcal{OMS}{cmsy}{b}{n}

\newcommand{\op}{\texttt{ETA-P}}
\newcommand{\stdp}{\texttt{STD-P}}

\begin{document}
\maketitle

\begin{abstract}
Despite recent advances in training and prompting strategies for Large Language Models (LLMs), these models continue to face challenges with complex logical reasoning tasks that involve long reasoning chains. In this work, we explore the potential and limitations of using graph-based synthetic reasoning data as training signals to enhance LLMs' reasoning capabilities. Our extensive experiments, conducted on two established natural language reasoning tasks—inductive reasoning and spatial reasoning—demonstrate that supervised fine-tuning (SFT) with synthetic graph-based reasoning data effectively enhances LLMs' reasoning performance, without compromising their effectiveness on other standard evaluation benchmarks. The code is publicly available at \url{https://github.com/riddickzhou/LLM-Graph-Synthetic-Reasoning}.
\end{abstract}

\section{Introduction}


The reasoning capabilities of Large Language Models (LLMs)~\cite{arxiv23_llama2,jiang2023Mistral7B,dubey2024llama} can be greatly enhanced by post-training techniques~\cite{ouyang2022training,zhang2023instruction} and prompting strategies~\cite{wei2022chain,yao2024tree,madaan2024self}.
However, even with the aforementioned techniques, 
the multi-hop reasoning tasks remain challenging~\cite{arxiv23_llama2,jiang2023Mistral7B,dubey2024llama}:
LLMs struggle to reason over steps~\cite{agrawal2024exploring,zhao2024exploring},
and are fragile to minor perturbations~\cite{ullman2023large,chen2024premise} in the input prompt.

Recently, several works~\cite{xu2023wizardlm, abdin2024phi, team2023gemini} have demonstrated the efficacy of boosting the LLMs' reasoning capacity via fine-tuning on synthetic data generated by stronger LLMs. 
However, how to make such synthetic data generation
effective and controllable for specific applications remains an open question. 
Extensive prompt engineering and quality filtering are required to guide LLMs' generation, yet the quality of generated reasoning questions and their labels remains uncertain~\cite{gudibande2023false, wang2023far, tan2024large}.

\begin{figure}[!h]
\includegraphics[width=0.75\linewidth]{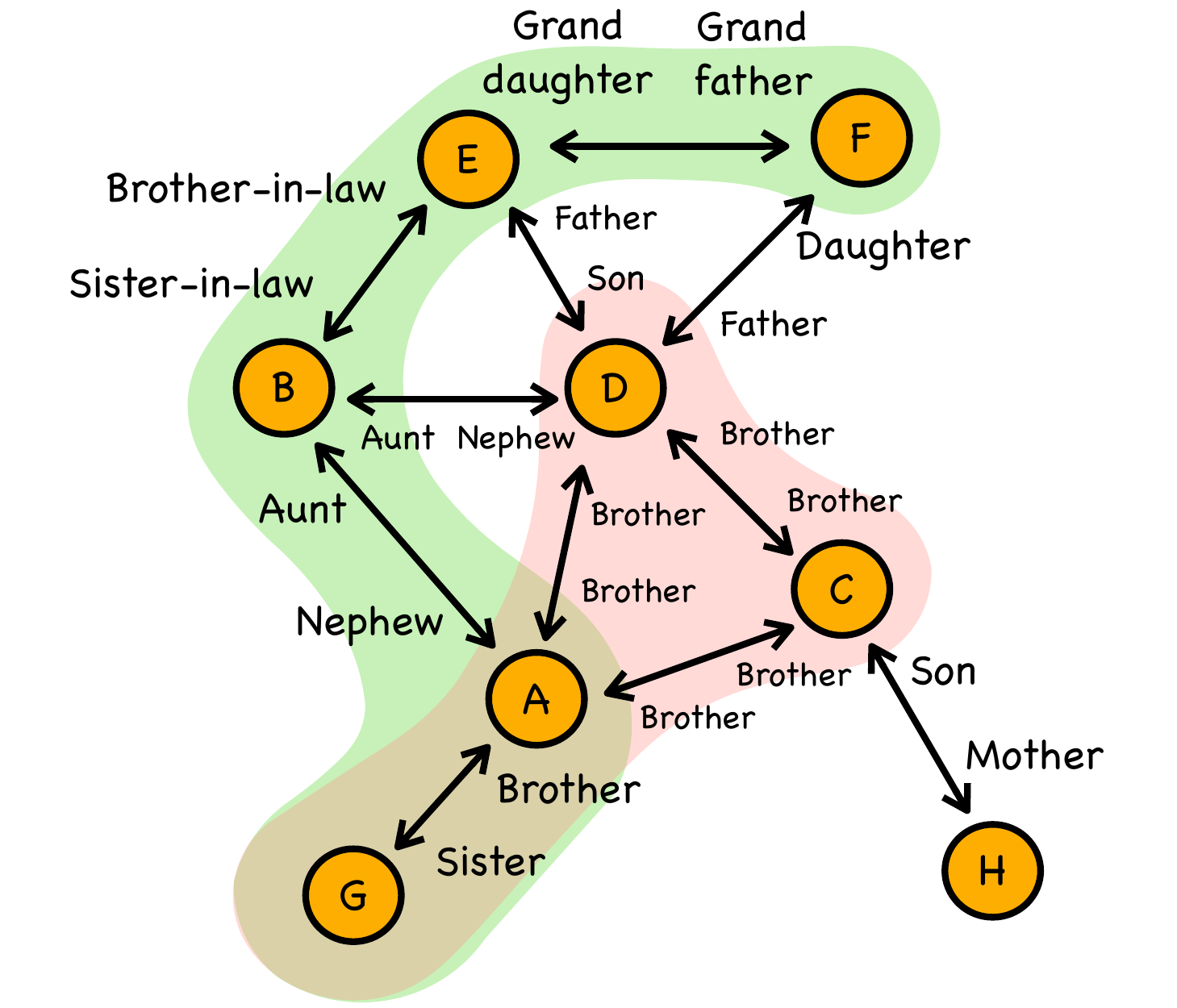}
\centering

\caption{Illustration of a kinship graph highlighting a reasoning chain sampled by our algorithm (green) for LLM adaptation, and an ignored simpler chain (red).}
\vspace{-3 mm}
\label{fig:main_figure}
\end{figure}

Motivated by the fact that natural language reasoning tasks can be represented as structured data with finite nodes and edges~\cite{jin2024graph}, and inspired by existing works on constructing reasoning benchmarks~\cite{fatemi2024TestTimeBenchmark, kazemi2023BoardgameQADatasetNatural, agrawal2024exploring},
we propose to leverage synthetic graph-based data for task-specific post-training adaptation
to improve the correctness of the generated reasoning questions and labels.

In this paper, we carefully design a random walk sampling algorithm on graphs and introduce a new prompting strategy that first extracts a reasoning chain and then derives the answer. These processes complement each other to enable efficient, task-specific adaptation of LLMs for reasoning tasks. Extensive experiments on two well-established benchmarks for inductive and spatial reasoning—\textit{CLUTRR}~\cite{sinha2019clutrr} and \textit{StepGame}~\cite{shi2022StepGameNewBenchmark}—demonstrate that our framework leads to significant performance gains compared to standard prompting and training methods. Our findings suggest that when carefully curated, graph-based structured data can effectively enhance LLMs' reasoning abilities on the targeted task while maintaining its performance on standard evaluation benchmarks. 

\color{blue}
\color{black}

\section{Methodology}
\label{sec:Methodology}


Reasoning tasks in natural language often involve a structured representation of facts that can be abstracted into a graph following predefined rules \cite{ji2022SurveyKnowledgeGraphs}. For example, family relationships can be systematically represented in a logical graph, where nodes denote family members and edges define their relationships, all governed by logical rules. Let \(\mathcal{G}{=}(\mathcal{V}, \mathcal{E}, \mathcal{R})\) represent a relational graph with a set of nodes \(\mathcal{V}\), a set of edges \(\mathcal{E}\) between the nodes, and a set of relations \(\mathcal{R}\) expressed in first-order logic. In this graph, vertices (e.g., family members) are denoted as \(v_i {\in} \mathcal{V}\), and directed edges (e.g., familial relationships) are represented as \((v_i, r_{i,j}, v_j) \in \mathcal{E}\), with  relations \(r_{i,j} {\in} \mathcal{R}\). Our goal is to generate synthetic examples of graph-structured data from such relation graphs to adapt LLMs for targeted reasoning tasks. Existing graph-based data generation methods, such as~\cite{sinha2019clutrr}, may produce large amounts of data, but this data is frequently redundant and lacks the necessary complexity (see \autoref{fig:main_figure}). Therefore, we propose a random-walk-based algorithm~\cite{lovasz1993random} that produces a manageable yet diverse set of examples by sampling sub-graphs from \(\mathcal{G}\). 
In the remainder of this section, we describe our algorithm for constructing $\mathcal{G}$ in~\S~\ref{sec:Relational Graph Construction}, the process of generating synthetic reasoning data in ~\S~\ref{sec:Sub-Graphs Sampling}, and their deployment to enhance LLM capabilities in natural language reasoning tasks in~\S~\ref{sec:Graph Synthetic Data for LLM Tuning}.

\subsection{Relational Graph Construction} 
\label{sec:Relational Graph Construction}

The relational graph $\gG$ is built iteratively by adding new nodes connected to existing nodes via basic relations.
We start with an initial graph $\gG_0 =(\{v_0\}, \emptyset, \cR)$, where $v_0$ is a randomly sampled root node. 
In each successive iteration $l$, we grow the graph by a) searching for absent relations between the nodes in $\gG_{l-1}$, and b) adding new nodes with those relations, if such absent relations are found in step a).
Specifically, for every node $v$ in $\gG_{l-1}$ and each relation $r \in \cR$, we check if there is another node $v^\prime$ in $\gG_{l-1}$ such that there is an edge between $v$ and $v^\prime$ with relation $r$. 
If no such node exists in $\gG_{l-1}$, we create a new node $v_r$ and connect it to node $v$ in $\gG_{l-1}$ with relation $r$. 
Then, a deduction function $f$ is used to evaluate the relation between $v_r$ and each other node $v'$ in $\gG_{l-1}$, except $v$. 
Edges $(v', r', v_r)$-s are added to $\gG_{l-1}$ if the deduction function $f$ computes a relation $r' \in \gR$ between $v'$ and $v_r$.
This process is repeated once for each node in $\gG_{l-1}$ within the $l$-th iteration.  
When this is complete, we assign the expanded $\gG_{l-1}$ to $\gG_{l}$ and proceed to the next iteration.
We terminate this procedure after $L$ iterations and take $\gG_{L}$ as the relational graph $\gG$. 



\subsection{Sub-Graphs Sampling}  \label{sec:Sub-Graphs Sampling}

Given a relational graph \(\mathcal{G}\) with $|\gV|{=}n$ nodes, the desired walking length (e.g., number of hops) $l {\in} \mathbb{Z}^+$, and isotropic random walk probabilities \(\pi {\in} \{0, 1\}^{n \times n}\) ($\pi_{i,j}$ denotes the probability of transition to $v_j$ from $v_i$ in one step of the random walk), we construct a reasoning chain $c$ by conducting a random walk of length $l$, which starts at a random node $v_0 {\in} \gV$ and avoids any repetition of visited nodes (to avoid circular reasoning in the data).
The overall procedure is summarized in Algorithm~\ref{alg:sg_sample}.

%

\begin{algorithm}[!thp]
\footnotesize 
\caption{Reasoning Chain Sampling}
\label{alg:sg_sample}
\textbf{Input}: Graph $\mathcal{G}$, walk length $l$
, transition probabilities $\pi$
\begin{algorithmic}[1]
\scriptsize
    \State Sample a starting vertex $v_0 \sim \text{Uniform}(\mathcal{V})$
    \State Initialize $walk \gets [v_0$]
    
    \For{$i = 0$ to $l-1$}
    \State $j \leftarrow i$
    \While{$v_j \in walk$}
    \State Sample $j \sim Categorical([\pi_{i,1}, \cdots \pi_{i,n}])$
    \If {$v_j \notin walk$}
    \State $v_{i+1} \leftarrow v_j$
    \State Append $v_{i+1}$ to $walk$ 
    \Break
     \EndIf
    \EndWhile
    \EndFor

    
    \State Initialize Reasoning chain $c \gets [ \ ]$
    \For{$i = 0$ to $l-1$}
        \State Retrieve and append $(v_i, r_{i,i+1}, v_{i+1}) \in \mathcal{E}$ to $c$ 
    \EndFor 
\end{algorithmic}
\textbf{Return}: Reasoning chain $c$
\vspace{-0.05cm}
\end{algorithm}
\vspace{-0.3 cm}

For each item \(c\) in the generated set of reasoning chains \(\mathcal{C}\), we apply one of the following augmentation techniques to further introduce diversity and additional complexity: \textbf{Permutation}, where we apply a permutation function \(\sigma\) to reorder the triples in \(c\); \textbf{Edge Noise}, which involves introducing noise by adding edges that connect nodes not initially in the chain, specifically, for a vertex \(v_i \in \mathcal{V}'\) from the chain, we add an edge to a vertex \(v_j \in \mathcal{V} \setminus \mathcal{V}'\): \((v_i, r_{i,j}, v_j)\); and \textbf{Edge Direction Flip} where we randomly flip the direction of some edges in \(c\), altering the flow of reasoning. 







\subsection{Graph Synthetic Data for LLM Tuning}  
\label{sec:Graph Synthetic Data for LLM Tuning}


The resultant set of reasoning chains \(\mathcal{C}\) can be converted into LLM-supervised fine-tuning data as follows. First, input-output pairs are created based on the requirements of the targeted reasoning task. For instance, a corruption function \( f^{\prime} \) operates by removing an edge \((v_i, r_{i,j}, v_j)\) from a chain \( c \), treating this edge as the output \( y \), while the remaining chain \( c' = c \setminus \{(v_i, r_{i,j}, v_j)\} \) is used as the input \( x \) for tasks such as family or spatial relation predictions. Second, the input \( c' \) needs to be converted into natural language textual input by applying a verbalizer, which can be either rule-based templates or more advanced techniques, such as utilizing a powerful LLM. Finally, while a standard prompt directs the model to answer immediately, we propose a graph-based reasoning task-specific prompting technique that mimics human cognitive processes in solving these types of tasks ~\cite{sinha2019clutrr}. We propose a new prompting technique we call 
\op{} (Extract then Answer) prompting, which instructs the model to extract the relational graph before generating the answer, as opposed to standard prompting \stdp{}, which instructs the LLM to directly generate the answer. Details of the prompt design are described in~\autoref{app:Prompt Design}. 

\color{blue}
\color{black}

\section{Experimental Setup}

\subsection{Tasks and Datasets}

To evaluate our framework, we benchmark it in two different logical reasoning tasks: ~\textbf{CLUTRR}~\cite{sinha2019clutrr} supports an inductive reasoning task that requires predicting the relationship between two family members (e.g. Alice is the sister of Bob) based on a story snippet that describes relevant familial relations;
\textbf{StepGame}~\cite{shi2022StepGameNewBenchmark} is a spatial reasoning task that involves determining the positional relationship between two entities (e.g., A is to the upper left of B) by navigating through a sequence of steps that describe relationships with neighboring entities. 
Both benchmarks feature logical reasoning problems with natural language story inputs followed by queries that require multi-hop reasoning. 
The accuracy of predicting the exact relationships is reported as the evaluation metric. Dataset statistics and synthetic data processing details are listed in Appendix~\ref{app:Data Processing}.

\subsection{Baselines}
We evaluate three system configurations, all using an instruction-tuned LLM as the backbone model.
\textbf{FS}: The model is tested in a few-shot setting with no additional tuning~\cite{brown2020LanguageModelsAre}; 
\textbf{SFT-S}: The model is supervised fine-tuned on the official training set of natural language stories; and \textbf{SFT-S}$+{k}$: The fine-tuning data consists of training story and $k$ systemic samples generated by our framework of ~\S~\ref{sec:Methodology}. For all main experiments\footnote{We also experiment with 2 other LLMs, see Appendix~\ref{app:Results}.}, we use Mistral-2-7B-Instruct~\cite{jiang2023Mistral7B} as the backbone LLM. Additionally, we include the few-shot test results of the commercial closed-source GPT-4o~\cite{openai2024hello} model. Implementation details of prompt design, fine-tuning, and inference hyperparameters can be found in Appendices~\ref{app:Baselines} and~\ref{app:Implementation Details}.


\section{Results and Analysis}

\subsection{Main Results}
\label{sec:Main Results}

\begin{figure}[!thp]
    \centering
    \begin{subfigure}{0.45\textwidth}
        \centering
        \includegraphics[width=\textwidth]{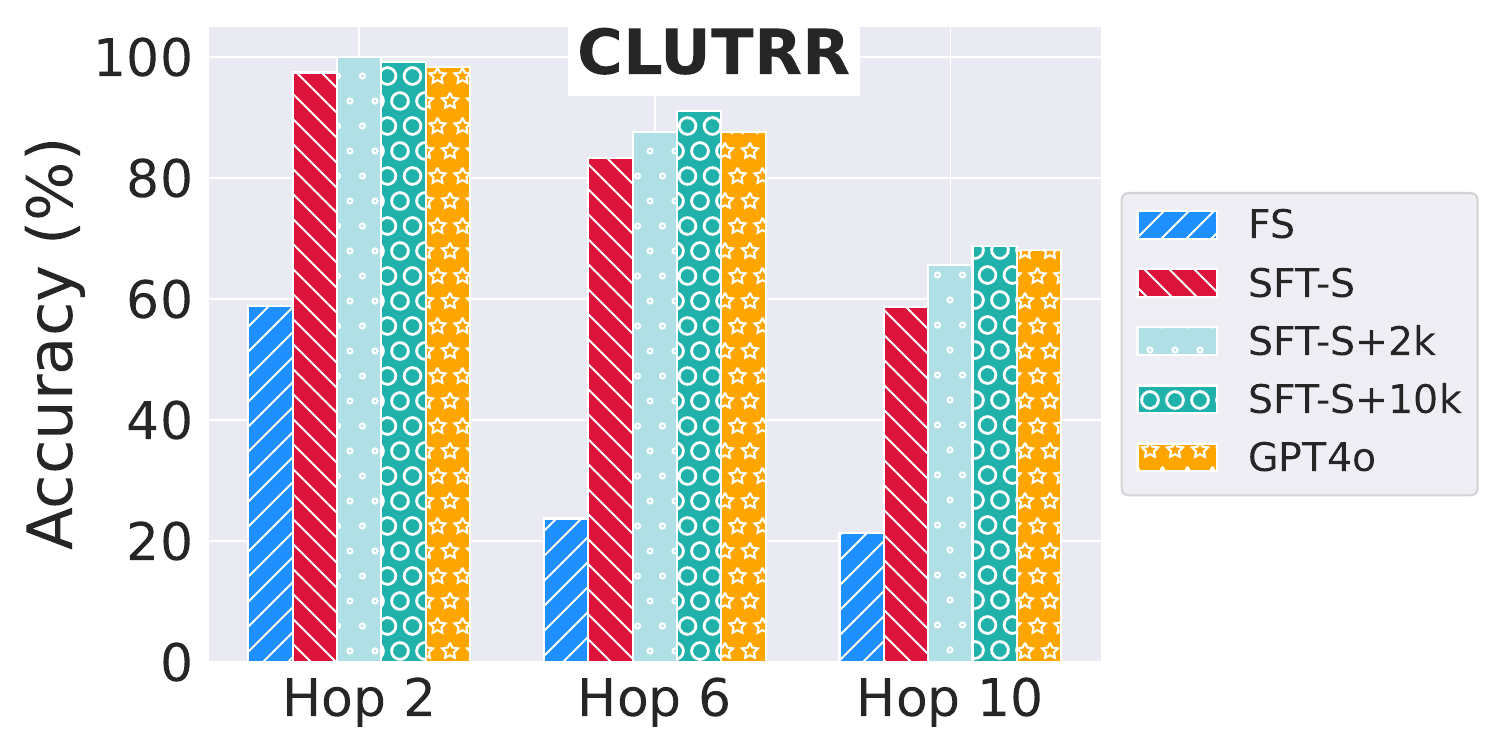}
    \end{subfigure}
    \begin{subfigure}{0.45\textwidth}
        \centering
        \includegraphics[width=\textwidth]{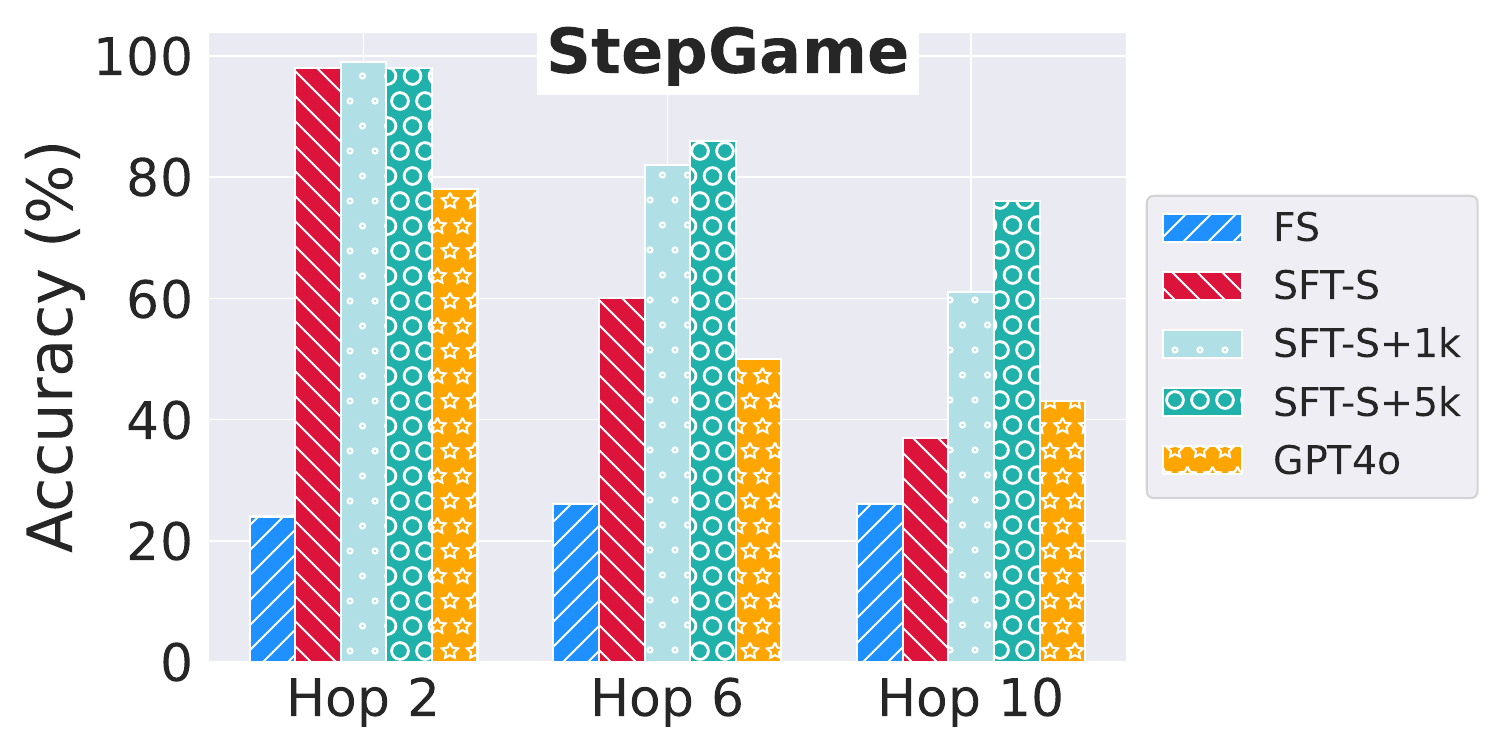}
    \end{subfigure}
    \caption{System performance on the CLUTRR (top) and StepGame (bottom) datasets for 2, 6, and 10 hop.}
    \label{fig:main_res}
    \vspace{-3mm}
\end{figure}

\autoref{fig:main_res} shows the performance on CLUTRR (top) and StepGame (bottom) of the Mistral-7B LLM under few-shot (\textbf{FS}), supervised fine-tuning on stories (\textbf{SFT-S}), and supervised fine-tuning on both stories and synthetic data of various sizes\footnote{The plot only shows the impact of adding the minimum and maximum amounts of synthetic data for CLUTRR ($+{2k}$ and $+{10k}$) and StepGame ($+{1k}$ and $+{5k}$).} (\textbf{SFT-S}$+{k}$) settings, on both the CLUTRR (top) and StepGame (bottom) datasets. In addition, it includes the few-shot performances of the GPT4-o model.\footnote{We were unable to include the results for all 10 hops and synthetic data size variants in the plot due to visualization constraints. Instead, we selected hops 2, 6, and 10 as representatives of the observed trends for low, mid, and high complexity reasoning, respectively. However, neighboring hops mostly exhibit similar result patterns, and detailed performance data are presented in Table~\ref{tab:main_res}
 in Appendix~\ref{app:Results}.}

First, we observe that on both datasets, \textbf{FS} significantly underperforms all \textbf{SFT} models across various reasoning hops, indicating that supervised fine-tuning is essential for enhancing performance on reasoning tasks in moderate-size open-source LLMs. Second, we observe that tuning with our synthetic data (\textbf{SFT-S}$+{k}$ models) consistently yields performance gains on the CLUTRR dataset, with these improvements becoming more pronounced at mid (6 hop) and high (10 hop) complexity levels of reasoning. Interestingly, we notice that extra tuning with synthetic data was necessary, as \textbf{SFT-S} underperformed compared to GPT-4o in most cases. This adaptation is crucial for open LLMs in domain-specific settings to achieve performance comparable to the GPT-4o model.

Nevertheless, we observe that on StepGame, a more challenging task with limited SFT data, training with our synthetic data leads to significant improvements compared to using only SFT on textual stories as well as when compared to the closed-source GPT-4o model across all hops. Furthermore, we notice that scaling up with our synthetic data results in more pronounced performance gaps between \textbf{SFT-S} and GPT-4o, compared to those observed on CLUTRR. These results suggest that synthetic data has a more significant impact on low-resource scenarios or challenging reasoning problems for task-specific adaptation of LLMs. Finally, we observe a systematic degradation in the performance of all models as we progress from low to mid to high hop reasoning complexity on both tasks, indicating that particularly complex reasoning cases continue to pose significant challenges for LLMs.

\subsection{Prompt Strategy Ablation}
\label{sec: Prompt Strategy  Ablation}

\begin{figure}[!thp]
    \centering
    \includegraphics[width=0.9\columnwidth]{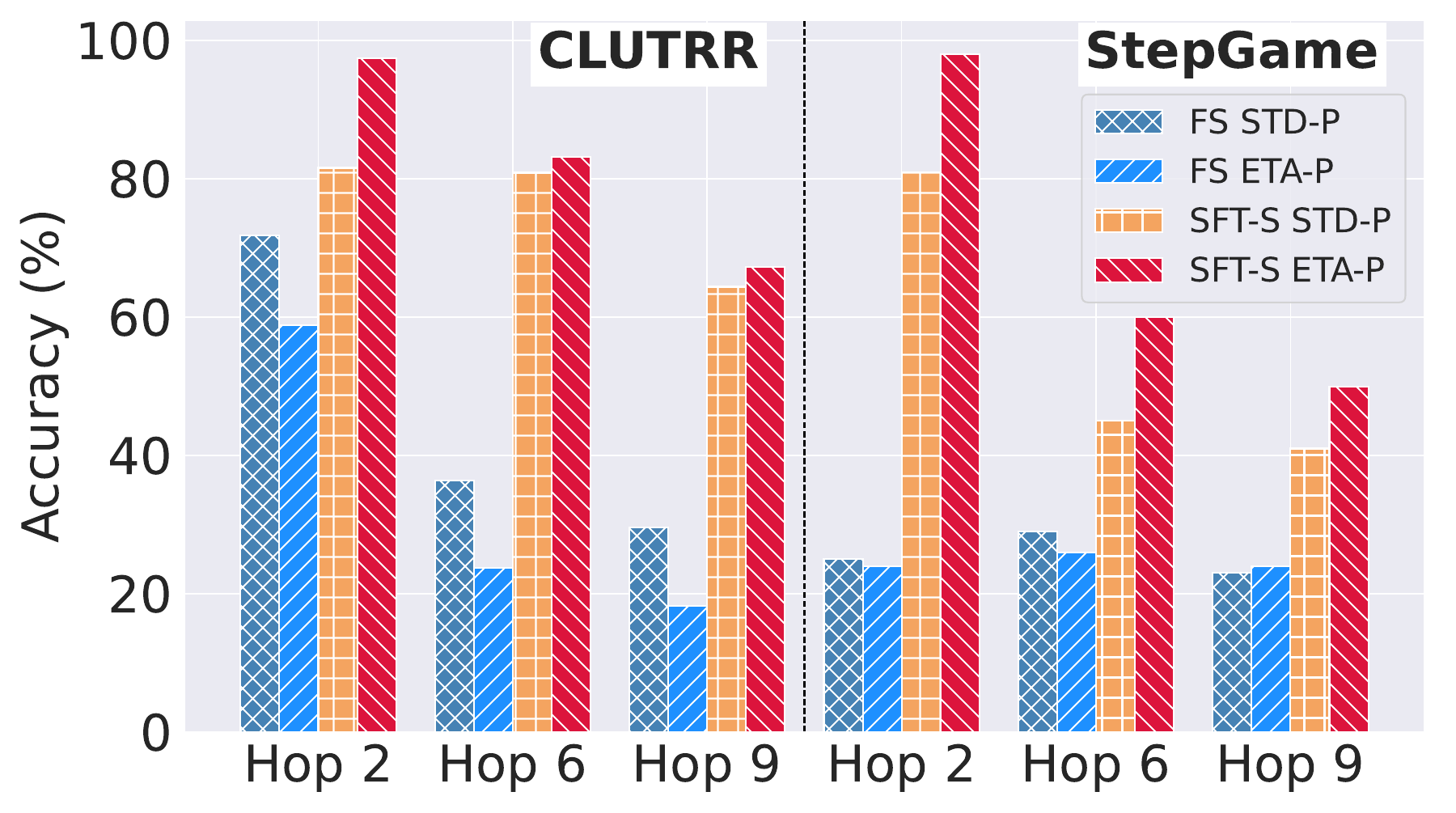}
    \caption{Mistral-2-7B performances on CLUTRR (left) and StepGame (right) datasets under \textbf{FS} and \textbf{SFT-S} settings when using \stdp{} and \op{} prompting.}
\label{fig:abl_prompt}
\vspace{-3mm}
\end{figure}

We study the impact of our proposed prompting strategy in \S~\ref{sec:Graph Synthetic Data for LLM Tuning} by comparing it to a model using standard prompting (see Table~\ref{tab:ablation_res} in~\S~\ref{app:Results}) in both few-shot (\textbf{FS}) and story-based supervised fine-tuning (\textbf{SFT-S}) settings; results are shown in \autoref{fig:abl_prompt}. 

Surprisingly, we observed that our task-specific prompt leads to performance drops under the \textbf{FS} setting\footnote{It is important to note that \textbf{FS} with \stdp{} consistently underperforms all \textbf{SFT} models across all settings.}. Our manual inspection revealed that models fail to correctly extract the graph relations from the story, resulting in error propagation into the predictions of the answer. In contrast, we notice that performing supervised fine-tuning (SFT) with our prompt leads to significant gains in most cases, compared to using the standard prompt during tuning. These results suggest that prompt engineering is complementary rather than a replacement for in-domain SFT for reasoning tasks.



\subsection{LLM Benchmarks Evaluation}

We validate whether LLMs can retain their open domain knowledge and problem-solving abilities, thereby avoiding catastrophic forgetting, after undergoing task-specific adaptation. We do so by evaluating models that have been tuned with the maximum amount of synthetic data, specifically \textbf{SFT-S}$+{10k}$ on CLUTRR and \textbf{SFT-S}$+{5k}$ on StepGame, on MMLU~\cite{hendrycksmeasuring}, GPQA~\cite{rein2023gpqa}, and GSM8K~\cite{cobbe2021training} benchmarks.

\begin{table}[h!]
\centering
\resizebox{\columnwidth}{!}{%
\begin{tabular}{l|ccc}
\toprule
\textbf{Benchmark} &\textbf{w.o. SFT} & \textbf{CLUTRR} & \textbf{StepGame }\\
\midrule
MMLU$_{ 0\textrm{-}shot}$& 57.54\%  & 58.46\% & 58.52\%\\
GPQA$_{ 0\textrm{-}shot}$& 29.46\%  & 29.02\% & 31.70\%\\
GSM8K$_{ 8\textrm{-}shot}$& 38.82\%  & 38.36\% & 38.74\%\\
\bottomrule
\end{tabular}
}

\caption{Performance of the original Mistral-2-7B, \textbf{SFT-S}$+{10k}$ on CLUTRR, and \textbf{SFT-S}$+{5k}$ on StepGame models across three LLM evaluation benchmarks.}
\label{tab:ben_eval}
\vspace{-3mm}
\end{table}

Results presented in~\autoref{tab:ben_eval} show minor variations in performance and, in some cases, improvements—as observed for \textbf{SFT-S}$+{10k}$ on CLUTTR during MMLU—across most benchmarks between the original model without SFT and those tuned on the two reasoning tasks. These observations suggest that task-specific adaptation of LLMs for reasoning tasks is feasible without sacrificing factual knowledge and generalization abilities, provided that the synthetic data for SFT is carefully curated.  



\section{Conclusion}
In this work, we propose a synthetic data augmentation algorithm and prompting strategy that effectively complement each other, enabling efficient task-specific adaptation of LLMs for reasoning tasks. We plan to expand our work to include a broader range of graph-based reasoning tasks. 

\section*{Limitations}
\color{blue}
\color{black}

Potential limitations of this work include not experimenting with larger open-source LLMs, such as the Mixtral 8x7B~\cite{jiang2024MixtralExperts} or 70B LLaMa3~\cite{llama3modelcard}. Additionally, the study did not incorporate more closed-source models such as Claude~\cite{claude3} to further study the impact of history enhancement. This is mainly due to limitations in computational (open source) and financial (closed source) resources.  While experiments show promise in enhancing LLM reasoning abilities without compromising their generalization performance, exploring more datasets and tasks can further enhance the credibility of our results.



\bibliography{ref, overleaf_citation_acl}

\begin{thebibliography}{40}
\providecommand{\natexlab}[1]{#1}

\bibitem[{Abdin et~al.(2024)Abdin, Jacobs, Awan, Aneja, Awadallah, Awadalla, Bach, Bahree, Bakhtiari, Behl et~al.}]{abdin2024phi}
Marah Abdin, Sam~Ade Jacobs, Ammar~Ahmad Awan, Jyoti Aneja, Ahmed Awadallah, Hany Awadalla, Nguyen Bach, Amit Bahree, Arash Bakhtiari, Harkirat Behl, et~al. 2024.
\newblock Phi-3 technical report: A highly capable language model locally on your phone.
\newblock \emph{arXiv preprint arXiv:2404.14219}.

\bibitem[{Agrawal et~al.(2024)Agrawal, Vasania, and Tan}]{agrawal2024exploring}
Palaash Agrawal, Shavak Vasania, and Cheston Tan. 2024.
\newblock Exploring the limitations of graph reasoning in large language models.
\newblock \emph{arXiv preprint arXiv:2402.01805}.

\bibitem[{AI@Meta(2024)}]{llama3modelcard}
AI@Meta. 2024.
\newblock Llama 3 model card.
\newblock \url{https://github.com/meta-llama/llama3/blob/main/MODEL_CARD.md}.

\bibitem[{Anil et~al.(2023)Anil, Borgeaud, Wu, Alayrac, Yu, Soricut, Schalkwyk, Dai, Hauth et~al.}]{team2023gemini}
Rohan Anil, Sebastian Borgeaud, Yonghui Wu, Jean-Baptiste Alayrac, Jiahui Yu, Radu Soricut, Johan Schalkwyk, Andrew~M Dai, Anja Hauth, et~al. 2023.
\newblock Gemini: a family of highly capable multimodal models.
\newblock \emph{arXiv preprint arXiv:2312.11805}.

\bibitem[{Anthropic(2024)}]{claude3}
Anthropic. 2024.
\newblock The claude 3 model family: Opus, sonnet, haiku.
\newblock \url{https://www-cdn.anthropic.com/de8ba9b01c9ab7cbabf5c33b80b7bbc618857627/Model_Card_Claude_3.pdf}.

\bibitem[{Brown et~al.(2020)Brown, Mann, Ryder, Subbiah, Kaplan, Dhariwal, Neelakantan, Shyam, Sastry, Askell, Agarwal, {Herbert-Voss}, Krueger, Henighan, Child, Ramesh, Ziegler, Wu, Winter, Hesse, Chen, Sigler, Litwin, Gray, Chess, Clark, Berner, McCandlish, Radford, Sutskever, and Amodei}]{brown2020LanguageModelsAre}
Tom~B. Brown, Benjamin Mann, Nick Ryder, Melanie Subbiah, Jared Kaplan, Prafulla Dhariwal, Arvind Neelakantan, Pranav Shyam, Girish Sastry, Amanda Askell, Sandhini Agarwal, Ariel {Herbert-Voss}, Gretchen Krueger, Tom Henighan, Rewon Child, Aditya Ramesh, Daniel~M. Ziegler, Jeffrey Wu, Clemens Winter, Christopher Hesse, Mark Chen, Eric Sigler, Mateusz Litwin, Scott Gray, Benjamin Chess, Jack Clark, Christopher Berner, Sam McCandlish, Alec Radford, Ilya Sutskever, and Dario Amodei. 2020.
\newblock \href {https://arxiv.org/abs/2005.14165} {Language {{Models}} are {{Few-Shot Learners}}}.

\bibitem[{Chen et~al.(2024)Chen, Chi, Wang, and Zhou}]{chen2024premise}
Xinyun Chen, Ryan~A Chi, Xuezhi Wang, and Denny Zhou. 2024.
\newblock Premise order matters in reasoning with large language models.
\newblock In \emph{Proc. Int. Conf. Mach. Learn.}

\bibitem[{Cobbe et~al.(2021)Cobbe, Kosaraju, Bavarian, Chen, Jun, Kaiser, Plappert, Tworek, Hilton, Nakano et~al.}]{cobbe2021training}
Karl Cobbe, Vineet Kosaraju, Mohammad Bavarian, Mark Chen, Heewoo Jun, Lukasz Kaiser, Matthias Plappert, Jerry Tworek, Jacob Hilton, Reiichiro Nakano, et~al. 2021.
\newblock Training verifiers to solve math word problems.
\newblock \emph{arXiv e-prints}, pages arXiv--2110.

\bibitem[{Dubey et~al.(2024)Dubey, Jauhri, Pandey, Kadian, Al-Dahle, Letman, Mathur, Schelten, Yang, Fan et~al.}]{dubey2024llama}
Abhimanyu Dubey, Abhinav Jauhri, Abhinav Pandey, Abhishek Kadian, Ahmad Al-Dahle, Aiesha Letman, Akhil Mathur, Alan Schelten, Amy Yang, Angela Fan, et~al. 2024.
\newblock The llama 3 herd of models.
\newblock \emph{arXiv preprint arXiv:2407.21783}.

\bibitem[{Fatemi et~al.(2024)Fatemi, Kazemi, Tsitsulin, Malkan, Yim, Palowitch, Seo, Halcrow, and Perozzi}]{fatemi2024TestTimeBenchmark}
Bahare Fatemi, Mehran Kazemi, Anton Tsitsulin, Karishma Malkan, Jinyeong Yim, John Palowitch, Sungyong Seo, Jonathan Halcrow, and Bryan Perozzi. 2024.
\newblock Test of {{Time}}: {{A Benchmark}} for {{Evaluating LLMs}} on {{Temporal Reasoning}}.
\newblock \emph{arXiv preprint arXiv:2406.09170}.

\bibitem[{Gudibande et~al.(2023)Gudibande, Wallace, Snell, Geng, Liu, Abbeel, Levine, and Song}]{gudibande2023false}
Arnav Gudibande, Eric Wallace, Charlie Snell, Xinyang Geng, Hao Liu, Pieter Abbeel, Sergey Levine, and Dawn Song. 2023.
\newblock The false promise of imitating proprietary llms.
\newblock \emph{arXiv preprint arXiv:2305.15717}.

\bibitem[{Hendrycks et~al.(2021)Hendrycks, Burns, Basart, Zou, Mazeika, Song, and Steinhardt}]{hendrycksmeasuring}
Dan Hendrycks, Collin Burns, Steven Basart, Andy Zou, Mantas Mazeika, Dawn Song, and Jacob Steinhardt. 2021.
\newblock Measuring massive multitask language understanding.
\newblock In \emph{Proc. Int. Conf. Learn. Represent.}

\bibitem[{Ji et~al.(2022)Ji, Pan, Cambria, Marttinen, and Yu}]{ji2022SurveyKnowledgeGraphs}
Shaoxiong Ji, Shirui Pan, Erik Cambria, Pekka Marttinen, and Philip~S. Yu. 2022.
\newblock A {{Survey}} on {{Knowledge Graphs}}: {{Representation}}, {{Acquisition}}, and {{Applications}}.
\newblock \emph{IEEE Transactions on Neural Networks and Learning Systems}, 33(2).

\bibitem[{Jiang et~al.(2023)Jiang, Sablayrolles, Mensch, Bamford, Chaplot, Casas, Bressand, Lengyel, Lample, Saulnier et~al.}]{jiang2023Mistral7B}
Albert~Q Jiang, Alexandre Sablayrolles, Arthur Mensch, Chris Bamford, Devendra~Singh Chaplot, Diego de~las Casas, Florian Bressand, Gianna Lengyel, Guillaume Lample, Lucile Saulnier, et~al. 2023.
\newblock Mistral 7b.
\newblock \emph{arXiv preprint arXiv:2310.06825}.

\bibitem[{Jiang et~al.(2024)Jiang, Sablayrolles, Roux, Mensch, Savary, Bamford, Chaplot, de~las Casas, Hanna, Bressand, Lengyel, Bour, Lample, Lavaud, Saulnier, Lachaux, Stock, Subramanian, Yang, Antoniak, Scao, Gervet, Lavril, Wang, Lacroix, and Sayed}]{jiang2024MixtralExperts}
Albert~Q. Jiang, Alexandre Sablayrolles, Antoine Roux, Arthur Mensch, Blanche Savary, Chris Bamford, Devendra~Singh Chaplot, Diego de~las Casas, Emma~Bou Hanna, Florian Bressand, Gianna Lengyel, Guillaume Bour, Guillaume Lample, L{\'e}lio~Renard Lavaud, Lucile Saulnier, Marie-Anne Lachaux, Pierre Stock, Sandeep Subramanian, Sophia Yang, Szymon Antoniak, Teven~Le Scao, Th{\'e}ophile Gervet, Thibaut Lavril, Thomas Wang, Timoth{\'e}e Lacroix, and William~El Sayed. 2024.
\newblock \href {https://arxiv.org/abs/2401.04088} {Mixtral of {{Experts}}}.

\bibitem[{Jin et~al.(2024)Jin, Xie, Zhang, Roy, Zhang, Wang, Meng, and Han}]{jin2024graph}
Bowen Jin, Chulin Xie, Jiawei Zhang, Kashob~Kumar Roy, Yu~Zhang, Suhang Wang, Yu~Meng, and Jiawei Han. 2024.
\newblock Graph chain-of-thought: Augmenting large language models by reasoning on graphs.
\newblock In \emph{Proc. Annu. Meet. Assoc. Comput. Linguist.}

\bibitem[{Kazemi et~al.(2023)Kazemi, Yuan, Bhatia, Kim, Xu, Imbrasaite, and Ramachandran}]{kazemi2023BoardgameQADatasetNatural}
Mehran Kazemi, Quan Yuan, Deepti Bhatia, Najoung Kim, Xin Xu, Vaiva Imbrasaite, and Deepak Ramachandran. 2023.
\newblock Boardgameqa: A dataset for natural language reasoning with contradictory information.
\newblock In \emph{Adv. Neural Inf. Process. Syst.}, volume~36.

\bibitem[{Loshchilov and Hutter(2017)}]{loshchilov2017decoupled}
Ilya Loshchilov and Frank Hutter. 2017.
\newblock Decoupled weight decay regularization.
\newblock \emph{arXiv preprint arXiv:1711.05101}.

\bibitem[{Lov{\'a}sz(1993)}]{lovasz1993random}
L{\'a}szl{\'o} Lov{\'a}sz. 1993.
\newblock Random walks on graphs.
\newblock \emph{Combinatorics, Paul erdos is eighty}, 2(1-46):4.

\bibitem[{Madaan et~al.(2023)Madaan, Tandon, Gupta, Hallinan, Gao, Wiegreffe, Alon, Dziri, Prabhumoye, Yang et~al.}]{madaan2024self}
Aman Madaan, Niket Tandon, Prakhar Gupta, Skyler Hallinan, Luyu Gao, Sarah Wiegreffe, Uri Alon, Nouha Dziri, Shrimai Prabhumoye, Yiming Yang, et~al. 2023.
\newblock Self-refine: Iterative refinement with self-feedback.
\newblock In \emph{Adv. Neural Inf. Process. Syst.}, volume~36.

\bibitem[{OpenAI(2024{\natexlab{a}})}]{openai2024hello}
OpenAI. 2024{\natexlab{a}}.
\newblock Hello {GPT-4o}.
\newblock \url{https://openai.com/index/hello-gpt-4o/}.
\newblock Accessed: 2024-05-26.

\bibitem[{OpenAI(2024{\natexlab{b}})}]{openai2024learning}
OpenAI. 2024{\natexlab{b}}.
\newblock \href {https://openai.com/index/learning-to-reason-with-llms/} {Learning to reason with llms}.
\newblock Accessed: 2024-09-19.

\bibitem[{Ouyang et~al.(2022)Ouyang, Wu, Jiang, Almeida, Wainwright, Mishkin, Zhang, Agarwal, Slama, Ray et~al.}]{ouyang2022training}
Long Ouyang, Jeffrey Wu, Xu~Jiang, Diogo Almeida, Carroll Wainwright, Pamela Mishkin, Chong Zhang, Sandhini Agarwal, Katarina Slama, Alex Ray, et~al. 2022.
\newblock Training language models to follow instructions with human feedback.
\newblock \emph{Advances in Neural Information Processing Systems}, 35:27730--27744.

\bibitem[{Paszke et~al.(2019)Paszke, Gross, Massa, Lerer, Bradbury, Chanan, Killeen, Lin, Gimelshein, Antiga et~al.}]{paszke2019pytorch}
Adam Paszke, Sam Gross, Francisco Massa, Adam Lerer, James Bradbury, Gregory Chanan, Trevor Killeen, Zeming Lin, Natalia Gimelshein, Luca Antiga, et~al. 2019.
\newblock Pytorch: An imperative style, high-performance deep learning library.
\newblock \emph{Advances in neural information processing systems}, 32:8026--8037.

\bibitem[{Rein et~al.(2023)Rein, Hou, Stickland, Petty, Pang, Dirani, Michael, and Bowman}]{rein2023gpqa}
David Rein, Betty~Li Hou, Asa~Cooper Stickland, Jackson Petty, Richard~Yuanzhe Pang, Julien Dirani, Julian Michael, and Samuel~R Bowman. 2023.
\newblock Gpqa: A graduate-level google-proof q\&a benchmark.
\newblock \emph{arXiv preprint arXiv:2311.12022}.

\bibitem[{Shi et~al.(2022)Shi, Zhang, and Lipani}]{shi2022StepGameNewBenchmark}
Zhengxiang Shi, Qiang Zhang, and Aldo Lipani. 2022.
\newblock {{StepGame}}: {{A New Benchmark}} for {{Robust Multi-Hop Spatial Reasoning}} in {{Texts}}.
\newblock In \emph{Proc. AAAI Conf. Artif. Intell.}, pages 11321--11329.

\bibitem[{Sinha et~al.(2019)Sinha, Sodhani, Dong, Pineau, and Hamilton}]{sinha2019clutrr}
Koustuv Sinha, Shagun Sodhani, Jin Dong, Joelle Pineau, and William~L Hamilton. 2019.
\newblock Clutrr: A diagnostic benchmark for inductive reasoning from text.
\newblock In \emph{Proc. Conf. Empir. Methods Nat. Lang. Process. Int. Joint Conf. Nat. Lang. Process.}, pages 4506--4515.

\bibitem[{Tan et~al.(2024)Tan, Beigi, Wang, Guo, Bhattacharjee, Jiang, Karami, Li, Cheng, and Liu}]{tan2024large}
Zhen Tan, Alimohammad Beigi, Song Wang, Ruocheng Guo, Amrita Bhattacharjee, Bohan Jiang, Mansooreh Karami, Jundong Li, Lu~Cheng, and Huan Liu. 2024.
\newblock Large language models for data annotation: A survey.
\newblock \emph{arXiv preprint arXiv:2402.13446}.

\bibitem[{Touvron et~al.(2023)Touvron, Martin, Stone, Albert, Almahairi, Babaei, Bashlykov, Batra, Bhargava, Bhosale et~al.}]{arxiv23_llama2}
Hugo Touvron, Louis Martin, Kevin Stone, Peter Albert, Amjad Almahairi, Yasmine Babaei, Nikolay Bashlykov, Soumya Batra, Prajjwal Bhargava, Shruti Bhosale, et~al. 2023.
\newblock Llama 2: Open foundation and fine-tuned chat models.
\newblock \emph{arXiv preprint arXiv:2307.09288}.

\bibitem[{Ullman(2023)}]{ullman2023large}
Tomer Ullman. 2023.
\newblock Large language models fail on trivial alterations to theory-of-mind tasks.
\newblock \emph{arXiv preprint arXiv:2302.08399}.

\bibitem[{Wang et~al.(2023)Wang, Ivison, Dasigi, Hessel, Khot, Chandu, Wadden, MacMillan, Smith, Beltagy et~al.}]{wang2023far}
Yizhong Wang, Hamish Ivison, Pradeep Dasigi, Jack Hessel, Tushar Khot, Khyathi Chandu, David Wadden, Kelsey MacMillan, Noah~A Smith, Iz~Beltagy, et~al. 2023.
\newblock How far can camels go? exploring the state of instruction tuning on open resources.
\newblock In \emph{Adv. Neural Inf. Process. Syst.}, volume~36, pages 74764--74786.

\bibitem[{Wei et~al.(2022{\natexlab{a}})Wei, Wang, Schuurmans, Bosma, Ichter, Xia, Chi, Le, and Zhou}]{wei2022chainThoughtPrompting}
Jason Wei, Xuezhi Wang, Dale Schuurmans, Maarten Bosma, Brian Ichter, Fei Xia, Ed~H. Chi, Quoc~V Le, and Denny Zhou. 2022{\natexlab{a}}.
\newblock Chain of thought prompting elicits reasoning in large language models.
\newblock In \emph{Adv. Neural Inf. Process. Syst.}

\bibitem[{Wei et~al.(2022{\natexlab{b}})Wei, Wang, Schuurmans, Bosma, Xia, Chi, Le, Zhou et~al.}]{wei2022chain}
Jason Wei, Xuezhi Wang, Dale Schuurmans, Maarten Bosma, Fei Xia, Ed~Chi, Quoc~V Le, Denny Zhou, et~al. 2022{\natexlab{b}}.
\newblock Chain-of-thought prompting elicits reasoning in large language models.
\newblock \emph{Advances in neural information processing systems}, 35:24824--24837.

\bibitem[{Wolf et~al.(2020)Wolf, Chaumond, Debut, Sanh, Delangue, Moi, Cistac, Funtowicz, Davison, Shleifer et~al.}]{wolf2020transformers}
Thomas Wolf, Julien Chaumond, Lysandre Debut, Victor Sanh, Clement Delangue, Anthony Moi, Pierric Cistac, Morgan Funtowicz, Joe Davison, Sam Shleifer, et~al. 2020.
\newblock Transformers: State-of-the-art natural language processing.
\newblock In \emph{Proceedings of the 2020 Conference on Empirical Methods in Natural Language Processing: System Demonstrations}, pages 38--45.

\bibitem[{Xu et~al.(2024)Xu, Sun, Zheng, Geng, Zhao, Feng, Tao, and Jiang}]{xu2023wizardlm}
Can Xu, Qingfeng Sun, Kai Zheng, Xiubo Geng, Pu~Zhao, Jiazhan Feng, Chongyang Tao, and Daxin Jiang. 2024.
\newblock Wizardlm: Empowering large language models to follow complex instructions.
\newblock In \emph{Int. Conf. Learn. Represent.}

\bibitem[{Yang et~al.(2024)Yang, Yang, Hui, Zheng, Yu, Zhou, Li, Li, Liu, Huang et~al.}]{qwen2}
An~Yang, Baosong Yang, Binyuan Hui, Bo~Zheng, Bowen Yu, Chang Zhou, Chengpeng Li, Chengyuan Li, Dayiheng Liu, Fei Huang, et~al. 2024.
\newblock Qwen2 technical report.
\newblock \emph{arXiv preprint arXiv:2407.10671}.

\bibitem[{Yang et~al.(2023)Yang, Ishay, and Lee}]{yang2023CouplingLargeLanguage}
Zhun Yang, Adam Ishay, and Joohyung Lee. 2023.
\newblock Coupling large language models with logic programming for robust and general reasoning from text.
\newblock In \emph{Findings of the Association for Computational Linguistics: ACL}.

\bibitem[{Yao et~al.(2023)Yao, Yu, Zhao, Shafran, Griffiths, Cao, and Narasimhan}]{yao2024tree}
Shunyu Yao, Dian Yu, Jeffrey Zhao, Izhak Shafran, Tom Griffiths, Yuan Cao, and Karthik Narasimhan. 2023.
\newblock Tree of thoughts: Deliberate problem solving with large language models.
\newblock In \emph{Adv. Neural Inf. Process. Syst.}

\bibitem[{Zhang et~al.(2023)Zhang, Dong, Li, Zhang, Sun, Wang, Li, Hu, Zhang, Wu et~al.}]{zhang2023instruction}
Shengyu Zhang, Linfeng Dong, Xiaoya Li, Sen Zhang, Xiaofei Sun, Shuhe Wang, Jiwei Li, Runyi Hu, Tianwei Zhang, Fei Wu, et~al. 2023.
\newblock Instruction tuning for large language models: A survey.
\newblock \emph{arXiv preprint arXiv:2308.10792}.

\bibitem[{Zhao and Zhang(2024)}]{zhao2024exploring}
Jinman Zhao and Xueyan Zhang. 2024.
\newblock Exploring the limitations of large language models in compositional relation reasoning.
\newblock \emph{arXiv preprint arXiv:2403.02615}.

\end{thebibliography}
\clearpage
\appendix

\section{Experimental Setup}

\subsection{Data Processing}
\label{app:Data Processing}
\autoref{tab:dataset_stat} shows the number of natural language story\footnote{In addition to the textual story, the structured data corresponding to each story is provided in each benchmark.} samples in the train and test splits of the CLUTRR~\cite{sinha2019clutrr} and StepGame~\cite{shi2022StepGameNewBenchmark} benchmarks per hop, as well as the total number of graph-based synthetic data we generate for each benchmark. 
Note that the original StepGame dataset consists of 
structured data for training and test splits only. 
We deployed GPT-4o to convert these splits into natural language story snippets for StepGame, thereby ensuring consistency in the experimental settings with the train-test splits of its counterpart in the CLUTRR benchmark. 

\begin{table}[!htp]
    \begin{center}

    \resizebox{\columnwidth}{!}{
\begin{tabular}{l|cc:c|cc:c}
\toprule
\multirow{2}{*}{Hop} & \multicolumn{3}{c|}{\textbf{CLUTRR}} & \multicolumn{3}{c}{\textbf{StepGame}} \\
& train & test & syn. & train & test & syn.\\
\midrule
2  & $333$  &  $114$   & $1162$    & $333$ & $100$ & $555$\\
3  & $352$  &  $229$   & $1170$    & $333$ & $100$ & $555$\\
4  & $317$  &  $219$   & $1129$    & $333$ & $100$ & $555$\\
5  & $-$    &  $308$   & $1219$    & $-$   & $100$ & $555$\\
6  & $-$    &  $178$   & $1224$    & $-$   & $100$ & $555$\\
7  & $-$    &  $246$   & $1231$    & $-$   & $100$ & $555$\\
8  & $-$    &  $228$   & $1120$    & $-$   & $100$ & $555$\\
9  & $-$    &  $172$   & $945$     & $-$   & $100$ & $555$\\
10 & $-$    &  $163$   & $795$     & $-$   & $100$ & $555$\\
\midrule
Total & $1002$ & $1857$ & $9995$ & $999$ & $900$ & $4995$\\
\bottomrule
\end{tabular}

    }
 \end{center}

\caption{Number of samples in the train-test splits of the CLUTRR and StepGame benchmarks per hop, as well as teh total number of synthetic data we populate for each benchmark.}
\label{tab:dataset_stat}
\end{table}

We use the algorithm described in~\S~\ref{sec:Methodology} in order to generate $10k$ and $5k$ reasoning chains for CLUTRR and StepGame, respectively. The synthetic samples are evenly distributed across the 9 different hop categories (2-10). We opted to generate twice as much synthetic data to support CLUTRR because its set of label classes is larger than that of StepGame.


We used the ASP solver from~\cite{yang2023CouplingLargeLanguage} as the deduction function $f$ in \S~\ref{sec:Relational Graph Construction} for CLUTRR deductive reasoning family relationship predict task. Conversely, for the StepGame spatial reasoning task, the relationship is deduced based on relative coordinates. For both tasks, the corruption function \( f' \) described in \S~\ref{sec:Relational Graph Construction} involves removing the edge between the head and tail of the reasoning chain \(c\) and using it as the output label. 
For both datasets, we employ a simple syntactic rule-based heuristic system as a verbalizer to convert our synthetic input reasoning chain \(c'\) into a natural language story. We prefer this cost-free verbalizer over LLM API options to accurately quantify the contribution of the synthetic chains themselves, without the enhancements provided by LLMs. Additionally, while LLM-based verbalizers may generate richer text stories, they can introduce errors in story generation, especially in larger hop scenarios. For each task, we carefully design both \stdp{} and \op{} prompts, resulting in a total of four prompts. The prompt design is described later in Appendix~\ref{app:Prompt Design}


\subsection{Baselines}
\label{app:Baselines}

For the Few-shot (\textbf{FS}) setting, we conduct experiments with Mistral-7B-Instruct and GPT-4o~\footnote{We include this model to benchmark our methods against a state-of-the-art closed-source system, which is treated as a measure to assess the upper bound of performance.} using both the \stdp{} and \op{} prompting strategies. In contrast, all settings involving fine-tuning (\textbf{SFT-S} and \textbf{SFT-S}\(+k\)) are conducted exclusively with Mistral-7B. For the \textbf{SFT-S} setting, we conduct experiments with both \stdp{} and \op{} settings, while we experiment only with \op{} for \textbf{SFT-S}\(+k\). For CLUTRR and StepGame, we create three configurations for \textbf{SFT-S}\(+k\) where \(k \in \{2000, 5000, 10000\}\) and \(k \in \{500, 2000, 5000\}\) respectively. These configurations simulate fine-tuning with small, medium, and large amounts of synthetic data. For the small and medium configurations, synthetic data are sampled proportionally across the 9 different hop categories.

In addition to Mistral-7B-Instruct, we conduct parallel experiments with two additional open-source models: Qwen2.5-7B-Instruct~\cite{qwen2} and Llama3-7B-Instruct~\cite{dubey2024llama}. Similar results and observed trends, as seen with Mistral-7B in Tables~\ref{tab:main_res},~\ref{tab:ablation_res}, and~\ref{tab:ben_eval}, are also observed with these two models in Tables~\ref{tab:main_res_qwen_llama},~\ref{tab:ablation_res_qwen_llama}, and~\ref{tab:ben_eval_qwen_llama}, respectively.

\subsection{Implementation Details}
\label{app:Implementation Details}

For \textbf{FS} experiments with GPT-4o, we access the model through the official OpenAI API~\footnote{We use the \texttt{gpt-4o-2024-05-13} version of the model from \url{https://chatgpt.com/}} using the default generation parameters. Inference and fine-tuning experiments were performed on a single GPU server that consists of 8 NVIDIA Tesla V100 cards with 32GB of memory. The pre-training code is based on the PyTorch~\cite{paszke2019pytorch} version of the Transformers library~\cite{wolf2020transformers}. In all fine-tuning experiments, we train Mistral-7B models for five epochs using a learning rate of 5e-7 with a batch size of 64. We always use AdamW~\cite{loshchilov2017decoupled} optimizer with a linear decay learning rate scheduler and a warm-up phase for the first 10\% of the training. During this phase, the learning rate gradually decreases to reach 1\% of its initial value at the end of the fine-tuning process. We found that five epochs were sufficient to fit fine-tuning data of all sizes, and the combinations of the learning rate and batch size were chosen to ensure numerical stability for each benchmark. During inference with all Mistral-7B models, we set the temperature to 0.01 and top-$k$ to 1 to minimize randomness during generation, consequently enhancing reproducibility.

\section{Prompt Design}
\label{app:Prompt Design}

In this section, we list the prompts that we have meticulously designed to tailor LLMs for specific reasoning tasks, including CLUTRR family relationship prediction and StepGame spatial relation reasoning. For each prompt, we designed the instruction part through trial and error iterations until we confirmed that both models (Mistral-7B and GPT-4o) could follow the instructions and generate outputs in the required format. In the few-shot setting, we set the number of in-context examples, that were picked up from training set of both datasets, to 5 as we did not see any improvement in adding more examples. We did not observe any benefits from using more in-context examples, as the outputs remained mostly stable, with minor to no changes in the model responses.

For both CLUTTER and StepGame, we design a standard (\stdp{}) and an 'extract then answer' \op{} prompt. As described in Section~\ref{sec:Graph Synthetic Data for LLM Tuning}, the  \stdp{} prompts the LLM to directly generate the answer, while \op{} requires first extracting the relational graph before producing the answer. Our proposed \op{} is similar to the step-by-step Chain of Thought (COT) prompting techniques~\cite{wei2022chainThoughtPrompting}, but it is specifically tailored for reasoning tasks, as it explicitly instructs the model to extract the relational graph before providing an answer. 

We do not position \op{} as an alternative or competitor to Chain-of-Thought (CoT) prompting for more general tasks. Rather, ETA-P is a specialized adaptation, which  aims to improve performance on graph-based reasoning tasks. While ETA-P can be viewed as a task-specific variant of CoT, it offers a key advantage in its ability to facilitate fine-tuning for graph-based tasks that require step-by-step reasoning. By leveraging synthetic triples as gold-standard intermediate steps, \op{} enables a more structured and targeted approach to reasoning, eliminating the need of collecting and annotating intermediate steps as is typically required when finetuning models with CoT-like reasoning strategy~\cite{openai2024learning}.

In our prompt, we use the following placeholders: \texttt{[STORY]} for the provided natural language story; \texttt{[QUERY]} for the natural language query; \texttt{[TRIPLES]} for the expected relational graph to extract in triplet format; \texttt{[ANSWER]} for the expected answer to be generated. Each of the prompts listed below has a few-shot version, which consists of placing five in-context examples in the same format (\#\#\# Story: ... \texttt{[ANSWER]}) just before the example of interest. Tables~\ref{tab:case1} to~\ref{tab:case6} show concrete examples of model prediction case analyses for illustrative purposes.

\subsection{CLUTRR \stdp{}}
\label{app:CLUTRR_stdp}
{\small
\texttt{You are given a narrative describing the familial relationships between several individuals. Analyze the narrative and determine the familial relationship between two specified individuals. The relationship between the characters must be the following: ['aunt', 'brother', 'daughter', 'daughter-in-law', 'father', 'father-in-law', 'granddaughter', 'grandfather', 'grandmother', 'grandson', 'mother', 'mother-in-law', 'nephew', 'niece', 'sister', 'son', 'son-in-law', 'uncle']} \\\\
\#\#\# Story: \\ \texttt{[STORY]}\\
\#\#\# Query: \\ \texttt{[Query]}\\\\
\#\#\# Output: \\ \texttt{[ANSWER]}
}

\subsection{CLUTRR \op{}}
\label{app:CLUTRR_op}
{\small
\texttt{You are given a narrative describing the familial relationships between several individuals. First break down the narrative into ordered structured triples, then attempt to answer the question. The relationship between the characters must be the following: ['aunt', 'brother', 'daughter', 'daughter-in-law', 'father', 'father-in-law', 'granddaughter', 'grandfather', 'grandmother', 'grandson', 'mother', 'mother-in-law', 'nephew', 'niece', 'sister', 'son', 'son-in-law', 'uncle']} \\
\#\#\# Story: \\ \texttt{[STORY]}\\
\#\#\# Query: \\ \texttt{[Query]}\\\\
\#\#\# Output: \\ The ordered structured triples are: \texttt{[TRIPLES]}. \\ Therefore, \texttt{[ANSWER]}
}

\subsection{StepGame \stdp{}}

{\small
\texttt{You are given a narrative describing the spatial relationships between several individuals. Analyze the narrative and determine the spatial relationship between two specified individuals. The relationship between the characters must be chosen from the following options: ["above", "below", "left", "lower-left", "lower-right", "right", "upper-left", "upper-right", "overlaps"]} \\\\
\#\#\# Story: \\ \texttt{[STORY]}\\
\#\#\# Query: \\ \texttt{[Query]}\\\\
\#\#\# Output: \\ \texttt{[ANSWER]}
}

\subsection{StepGame \op{}}
\label{app:StepGame_op}

{\small
\texttt{Prompt:\\
You are given a narrative describing the spatial relationships between several individuals. First break down the narrative into ordered structured triples, then attempt to answer the question. The relationship between the characters must be the following: ["above", "below", "left", "lower-left", "lower-right", "right", "upper-left", "upper-right", "overlaps"]} \\
\#\#\# Story: \\ \texttt{[STORY]}\\
\#\#\# Query: \\ \texttt{[Query]}\\\\
\#\#\# Output: \\ The ordered structured triples are: \texttt{[TRIPLES]}. \\ Therefore, \texttt{[ANSWER]}
}

\section{Results}
\label{app:Results}

\begin{table*}[!thp]
\resizebox{\textwidth}{!}{
\centering
\begin{tabular}{c|cccccc|cccccc}
\toprule

\multirow{3}{*}{Hop}& \multicolumn{6}{c|}{{\bf CLUTRR}} & \multicolumn{6}{c}{{\bf StepGame}}\\
& \multicolumn{5}{c}{{\bf  Mistral-7B-Instruct}}
& \bf GPT-4o
& \multicolumn{5}{c}{{\bf  Mistral-7B-Instruct}}
& \bf GPT-4o
\\
\cmidrule(lr){1-1} \cmidrule(lr){2-6} \cmidrule(lr){7-7} \cmidrule(lr){8-12} \cmidrule(lr){13-13}
& \footnotesize \textbf{FS} & \footnotesize \textbf{SFT-S} & \footnotesize \textbf{SFT-S$+2k$} & \footnotesize \textbf{SFT-S$+5k$} & \footnotesize \textbf{SFT-S$+10k$} & \footnotesize \textbf{FS}
& \footnotesize \textbf{FS} & \footnotesize \textbf{SFT-S} & \footnotesize \textbf{SFT-S$+1k$} & \footnotesize \textbf{SFT-S$+2k$} & \footnotesize \textbf{SFT-S$+5k$} & \footnotesize \textbf{FS} \\
$2$  & 58.8 & 97.4 & 100.0 & 100.0 & 99.1 & 98.3 & 24.0 & 98.0 & 99.0 & 100.0 & 98.0 & 78.0 \\
$3$  & 15.3 & 93.4 & 96.4 & 95.3 & 95.7 & 97.1 & 26.0 & 90.0 & 95.0 & 96.0 & 97.0 & 47.0  \\
$4$  & 33.2 & 84.4 & 87.2 & 88.1 & 88.1 & 85.8 & 27.0 & 74.0 & 90.0 & 92.0 & 92.0 & 45.0  \\
$5$  & 25.8 & 86.1 & 88.0 & 90.9 & 91.9 & 88.0 & 24.0 & 76.0 & 87.0 & 94.0 & 95.0 & 54.0  \\
$6$  & 23.7 & 83.2 & 87.6 & 87.6 & 91.0 & 87.6 & 26.0 & 60.0 & 82.0 & 91.0 & 87.0 & 50.0  \\
$7$  & 20.1 & 72.8 & 82.1 & 79.7 & 83.7 & 79.3 & 17.0 & 59.0 & 73.0 & 82.0 & 83.0 & 50.0  \\
$8$  & 21.8 & 68.6 & 76.7 & 77.6 & 78.5 & 75.0 & 12.0 & 48.0 & 75.0 & 76.0 & 79.0 & 39.0  \\
$9$  & 18.2 & 67.3 & 69.6 & 74.4 & 74.4 & 73.8 & 24.0 & 50.0 & 66.0 & 76.0 & 80.0 & 41.0  \\
$10$ & 21.3 & 58.6 & 65.6 & 64.4 & 68.7 & 68.1 & 26.0 & 37.0 & 61.0 & 78.0 & 76.0 & 43.0  \\
\bottomrule
\end{tabular}
}
\caption{Performance, in terms of accuracy, on the CLUTRR and StepGame benchmarks for the few-shot setting (\textbf{FS}) with the GPT-4o model, as well as the Mistral-7B-Instruct under \textbf{FS}, story supervised fine-tuning (\textbf{SFT-S}), and when using our synthetic data (\textbf{SFT-S$+k$}). We report the performance using when \op{} for all models.
}
\label{tab:main_res}
\end{table*}

\begin{table*}[!thp]
\resizebox{\textwidth}{!}{
\centering
\begin{tabular}{cc|ccccc|ccccc}
\toprule
& \bf Hop & \multicolumn{5}{c|}{{\bf Qwen2.5-7B-instruct}}
& \multicolumn{5}{c}{{\bf  Llama3-8B-Instruct}}
\\
\cmidrule(lr){2-2} \cmidrule(lr){3-7} \cmidrule(lr){8-12}
& & \footnotesize \textbf{FS} & \footnotesize \textbf{SFT-S} & \footnotesize \textbf{SFT-S$+2k$} & \footnotesize \textbf{SFT-S$+5k$} & \footnotesize \textbf{SFT-S$+10k$}
& \footnotesize \textbf{FS} & \footnotesize \textbf{SFT-S} & \footnotesize \textbf{SFT-S$+2k$} & \footnotesize \textbf{SFT-S$+5k$} & \footnotesize \textbf{SFT-S$+10k$}  \\
\multicolumn{1}{l}{\multirow{9}{*}{\begin{sideways}\specialcell{\bf CLUTRR}\end{sideways}}} & $2$  &  52.6 &  99.1 & 100.0 &  100.0 &  98.3 &  44.7 & 71.9 & 92.9 & 89.5 & 88.6 \\
& $3$  &  45.4 &  83.0 &  96.4 &  93.5 &  97.1 &  32.8 & 59.0 & 82.1 & 91.7 & 95.2 \\
& $4$  &  52.1 &  65.8 &  87.2 &  83.1 & 84.9 &  36.1 & 60.7 & 69.4 & 66.7 & 73.0 \\
& $5$  &  55.5 &  63.0 &  88.0 &  80.5 & 83.4 &  37.3 & 56.8 & 64.0 & 64.3 & 70.8 \\
& $6$  &  59.6 &  65.7 &  87.6 &  81.5 & 82.6 &  43.3 & 69.1 & 72.5 & 72.5 & 73.0 \\
& $7$  &  52.0 &  61.4 &  82.1 &  72.0 & 70.3 &  39.8 & 60.6 & 65.5 & 65.5 & 72.8 \\
& $8$  &  52.6 &  55.7 &  76.7 &  73.7 & 70.6 &  32.9 & 57.5 & 64.9 & 66.2 & 64.5 \\
& $9$  &  45.9 &  49.4 &  69.6 &  64.0 & 57.0 &  34.3 & 58.7 & 58.3 & 58.1 & 58.1 \\
& $10$ &  54.6 &  57.1 &  65.6 &  63.2 & 54.0 &  34.4 & 61.4 & 58.3 & 52.8 & 61.9 \\
\midrule
\multicolumn{1}{l}{\multirow{9}{*}{\begin{sideways}\specialcell{\bf StepGame}\end{sideways}}} & $2$ & 15.0 & 40.0 & 64.0 & 81.0 & 82.0 & 21.0 & 42.0 & 72.0 & 80.0 & 84.0 \\
& $3$ &  21.0 & 29.0 & 44.0 & 54.0 & 69.0 & 21.0 & 34.0 & 50.0 & 75.0 & 76.0 \\
& $4$ & 21.0 & 23.0 & 36.0 & 39.0 & 42.0 & 21.0 & 31.0 & 32.0 & 56.0 & 57.0 \\
& $5$ & 20.0 & 31.0 & 44.0 & 42.0 & 50.0 & 15.0 & 28.0 & 42.0 & 59.0 & 65.0 \\
& $6$ & 21.0 & 18.0 & 32.0 & 31.0 & 31.0 & 13.0 & 29.0 & 39.0 & 55.0 & 50.0 \\
& $7$ & 16.0 & 16.0 & 33.0 & 40.0 & 33.0 & 12.0 & 28.0 & 40.0 & 47.0 & 48.0 \\
& $8$ & 13.0 & 15.0 & 20.0 & 35.0 & 38.0 & 9.0  & 26.0 & 36.0 & 51.0 & 48.0 \\
& $9$ & 18.0 & 15.0 & 24.0 & 37.0 & 25.0 & 7.0  & 23.0 & 29.0 & 44.0 & 45.0  \\
& $10$ & 18.0 & 22.0 & 26.0 & 31.0 & 33.0 & 9.0  & 22.0 & 30.0 & 34.0 & 39.0 \\
\bottomrule

\end{tabular}
}

\caption{Performance, in terms of accuracy, on the CLUTRR and StepGame benchmarks under few-shot setting (\textbf{FS}), story supervised fine-tuning (\textbf{SFT-S}), and when using our synthetic data (\textbf{SFT-S$+k$}) for both Qwen2.5-7B-instruct and Llama3-8B-Instruct. We report the performance using when \op{} for both models.}
\label{tab:main_res_qwen_llama}
\end{table*}
\clearpage

\begin{table*}[!thp]
\resizebox{\textwidth}{!}{
\centering
\begin{tabular}{c|cccccc|cccccc}
\toprule
\multirow{3}{*}{Hop} & \multicolumn{6}{c|}{{\bf CLUTRR}} & \multicolumn{6}{c}{{\bf StepGame}}\\
& \multicolumn{2}{c}{{\bf Mistral FS}}
& \multicolumn{2}{c}{{\bf Mistral SFT-S}}
& \multicolumn{2}{c|}{{\bf GPT-4o FS}}
& \multicolumn{2}{c}{{\bf Mistral FS}}
& \multicolumn{2}{c}{{\bf Mistral SFT-S}}
& \multicolumn{2}{c}{{\bf GPT-4o FS}}
\\
\cmidrule(lr){1-1} \cmidrule(lr){2-3} \cmidrule(lr){4-5} \cmidrule(lr){6-7} \cmidrule(lr){8-9} \cmidrule(lr){10-11} \cmidrule(lr){12-13}
& \textbf{\stdp{}} &  \textbf{\op{}} & \textbf{\stdp{}} &  \textbf{\op{}} & \textbf{\stdp{}} &  \textbf{\op{}} & \textbf{\stdp{}} &  \textbf{\op{}} & \textbf{\stdp{}} &  \textbf{\op{}} & \textbf{\stdp{}} &  \textbf{\op{}}   \\
$2$  & 71.8 & 58.8 & 81.6 & 97.4 & 90.4 & 98.3 & 25.0 & 24.0 & 81.0 & 98.0 & 55.0 & 78.0 \\
$3$  & 14.7 & 15.3 & 93.0 & 93.4 & 95.1 & 97.1 & 28.0 & 26.0 & 66.0 & 90.0 & 52.0 & 47.0 \\
$4$  & 30.3 & 33.2 & 80.2 & 84.4 & 80.4 & 85.8 & 26.0 & 27.0 & 60.0 & 74.0 & 49.0 & 45.0 \\
$5$  & 37.2 & 25.8 & 76.1 & 86.1 & 71.8 & 88.0 & 29.0 & 24.0 & 62.0 & 76.0 & 51.0 & 54.0 \\
$6$  & 36.4 & 23.7 & 80.9 & 83.2 & 68.5 & 87.6 & 29.0 & 26.0 & 45.0 & 60.0 & 46.0 & 50.0 \\
$7$  & 35.7 & 20.1 & 63.0 & 72.8 & 64.6 & 79.3 & 31.0 & 17.0 & 48.0 & 59.0 & 47.0 & 50.0 \\
$8$  & 34.5 & 21.8 & 66.8 & 68.6 & 55.7 & 75.0 & 23.0 & 12.0 & 43.0 & 48.0 & 45.0 & 39.0 \\
$9$  & 29.6 & 18.2 & 64.4 & 67.3 & 50.6 & 73.8 & 23.0 & 24.0 & 41.0 & 50.0 & 47.0 & 41.0 \\
$10$ & 35.3 & 21.3 & 65.1 & 58.6 & 54.0 & 68.1 & 31.0 & 26.0 & 40.0 & 37.0 & 45.0 & 43.0 \\
\bottomrule
\end{tabular}
}

\caption{Performance, in terms of accuracy, on the CLUTRR and StepGame benchmarks for 3 models: {{\bf GPT-4o FS}}, {{\bf  Mistral-7B-Instruct FS}}, and {{\bf  Mistral-7B-Instruct SFT-S}}. For each model, we ablate when using standard (\stdp{}) and our extract then answer (\op{}) prompting strategies. 
}

\label{tab:ablation_res}
\end{table*}

\begin{table*}[!thp]
\centering
\begin{tabular}{cc|cccccccc}
\toprule
& Hop & \multicolumn{2}{c}{{\bf Qwen2.5 FS}}
& \multicolumn{2}{c}{{\bf Qwen2.5 SFT-S}}
& \multicolumn{2}{c}{{\bf Llama3 FS}}
& \multicolumn{2}{c}{{\bf Llama3 SFT-S}} 
\\
\cmidrule(lr){2-2} \cmidrule(lr){3-4} \cmidrule(lr){5-6} \cmidrule(lr){7-8} \cmidrule(lr){9-10} 
& & \textbf{\stdp{}} &  \textbf{\op{}} & \textbf{\stdp{}} &  \textbf{\op{}} & \textbf{\stdp{}} &  \textbf{\op{}} & \textbf{\stdp{}} &  \textbf{\op{}}  \\
\multicolumn{1}{l}{\multirow{9}{*}{\begin{sideways}\specialcell{\bf CLUTRR}\end{sideways}}} & $2$  & 80.7 & 52.6 & 99.1 & 99.1 & 67.5 & 44.7 & 89.5 & 71.9 \\
& $3$  & 61.1 & 45.4 & 66.4 & 83.0 & 38.0 & 32.8 & 89.1 & 59.0 \\
& $4$  & 54.3 & 52.1 & 61.2 & 65.8 & 37.9 & 36.1 & 69.4 & 60.7 \\
& $5$  & 52.9 & 55.5 & 59.1 & 63.0 & 30.5 & 37.3 & 65.6 & 56.8 \\
& $6$  & 54.5 & 59.6 & 56.2 & 65.7 & 27.5 & 43.3 & 59.6 & 69.1 \\
& $7$  & 54.0 & 52.0 & 52.0 & 61.4 & 23.6 & 39.8 & 63.0 & 60.6 \\
& $8$  & 54.0 & 52.6 & 50.4 & 55.7 & 24.1 & 32.9 & 54.8 & 57.5 \\
& $9$  & 47.7 & 45.9 & 46.5 & 49.4 & 21.5 & 34.3 & 47.7 & 58.7 \\
& $10$ & 57.1 & 54.6 & 50.9 & 57.0 & 23.3 & 34.4 & 50.9 & 61.4 \\
\midrule
\multicolumn{1}{l}{\multirow{9}{*}{\begin{sideways}\specialcell{\bf StepGame}\end{sideways}}} & $2$  & 17.0 & 15.0 & 17.0 & 40.0 & 16.0 & 21.0 & 37.0 & 42.0 \\
& $3$  & 23.0 & 21.0 & 17.0 & 29.0 & 17.0 & 21.0 & 43.0 & 34.0  \\
& $4$  & 16.0 & 21.0 & 17.0 & 23.0 & 16.0 & 21.0 & 40.0 & 31.0 \\
& $5$  & 11.0 & 20.0 & 18.0 & 31.0 & 14.0 & 15.0 & 45.0 & 28.0 \\
& $6$  & 17.0 & 21.0 & 6.0 & 18.0 & 10.0 & 13.0 & 41.0 & 29.0 \\
& $7$  & 9.0 & 16.0 & 8.0 & 16.0 & 12.0 & 12.0 & 38.0 & 28.0 \\
& $8$  & 13.0 & 13.0 & 15.0 & 15.0 & 10.0 & 9.0 & 30.0 & 26.0 \\
& $9$  & 17.0 & 18.0 & 12.0 & 15.0 & 10.0 & 7.0 & 36.0 & 23.0 \\
& $10$ & 12.0 & 18.0 & 7.0 & 22.0 & 10.0 & 9.0 & 27.0 & 22.0 \\

\bottomrule
\end{tabular}

\caption{Performance, in terms of accuracy, on the CLUTRR and StepGame benchmarks for {{\bf Qwen2.5-7B-Instruct FS}}, {{\bf Qwen2.5-7B-Instruct SFT-S}}, {{\bf Llama3-8B-Instruct FS}}, and {{\bf Llama3-8B-Instruct SFT-S}}. For each model, we ablate when using standard (\stdp{}) and our extract-then-answer (\op{}) prompting strategies. 
}
\label{tab:ablation_res_qwen_llama}
\end{table*}


\begin{table*}[h!]
\centering
\begin{tabular}{l|ccc|ccc}
\toprule
\multirow{2}{*}{\textbf{Benchmark}} & \multicolumn{3}{c|}{{\bf Qwen2.5-7B-Instruct}} & \multicolumn{3}{c}{{\bf Llama3-7B-Instruct}}\\
 &\textbf{w.o. SFT} & \textbf{CLUTRR} & \textbf{StepGame} &\textbf{w.o. SFT} & \textbf{CLUTRR} & \textbf{StepGame}\\
\midrule
MMLU$_{ 0\textrm{-}shot}$& 17.64\%  & 24.56\% & 25.47\% & 61.66\%  & 63.42\% & 59.63\%\\
GPQA$_{ 0\textrm{-}shot}$& 32.14\%  & 33.26\% & 33.48\% & 31.03\%  & 29.46\% & 28.35\%\\
GSM8K$_{ 8\textrm{-}shot}$& 69.29\%  & 71.34\% & 71.424\% & 76.35\%  & 74.98\% & 76.8\%\\
\bottomrule
\end{tabular}

\caption{Performance comparison of the Qwen2.5-7B-Instruct and Llama3-7B-Instruct models in their original form, \textbf{SFT-S}$+{10k}$ fine-tuning on CLUTRR, and \textbf{SFT-S}$+{5k}$ fine-tuning on StepGame, evaluated across three benchmarks.}
\label{tab:ben_eval_qwen_llama}
\end{table*}

\begin{table*}[t]
    \centering
    \resizebox{\textwidth}{!}{
    \begin{tabular}{p{0.5\textwidth}|p{0.6\textwidth}}
    \toprule
    \textbf{Prompt~\ref{app:CLUTRR_op} Placeholder Value} & \textbf{Prediction} \\
    \midrule
    \texttt{[STORY]} \newline
    Morgan was making dinner when her daughter Frances walked into the kitchen. Frances asked her brother Donald if he wanted to help set the table. Donald's sister Danielle was busy finishing her homework. Kevin, Danielle's grandfather, was reading a newspaper in the living room. Valerie, Kevin's daughter, was knitting a scarf. Harry, Valerie's brother, was watching a football game on TV. Kenneth, Harry's brother, called from the other room asking for help with a project. Kenneth's son Robert was busy working on his science project. Elizabeth, Robert's sister, was reading a book. Brittney, Elizabeth's sister, was playing a board game.\newline

    \texttt{[QUERY]}\newline
    What is the relationship of Brittney to Morgan?\newline

    \texttt{[TRIPLES]}\newline
    Brittney is the sister of Elizabeth.\newline
    Elizabeth is the sister of Robert.\newline
    Robert is the son of Kenneth.\newline
    Kenneth is the brother of Harry.\newline
    Harry is the brother of Valerie.\newline
    Valerie is the daughter of Kevin.\newline
    Kevin is the grandfather of Danielle.\newline
    Danielle is the sister of Donald.\newline
    Donald is the brother of Frances.\newline
    Frances is the daughter of Morgan.\newline

    \texttt{[ANSWER]} \newline
    Brittney is the \underline{niece} of Morgan \newline
    &

    \textbf{GPT-4o:} \newline
    The ordered structured triples are:\newline
    - Frances is the daughter of Morgan.\newline
    - Donald is the brother of Frances.\newline
    - Danielle is the sister of Donald.\newline
    - Kevin is the grandfather of Danielle.\newline
    - Valerie is the daughter of Kevin.\newline
    - Harry is the brother of Valerie.\newline
    - Kenneth is the brother of Harry.\newline
    - Robert is the son of Kenneth.\newline
    - Elizabeth is the sister of Robert.\newline
    - Brittney is the sister of Elizabeth.\newline

    Analyzing the relationships:\newline
    - Frances, Donald, and Danielle are siblings and children of Morgan.\newline
    - Kevin is the grandfather of Danielle, which means Kevin is the father of one of Morgan's parents.\newline
    - Valerie is the daughter of Kevin, making her Morgan's mother.\newline
    - Harry and Kenneth are Valerie's brothers, making them Morgan's uncles.\newline
    - Robert, Elizabeth, and Brittney are Kenneth's children, making them Morgan's cousins.\newline

    Therefore,
    Brittney is the \underline{\textcolor{red}{cousin}} of Morgan.\newline

    \textbf{SFT-S}: \newline
    The ordered structured triples are: \newline
    Brittney is the granddaughter of Kevin, who is the father of Valerie, who is the sister of Harry, who is the brother of Kenneth, whose son Robert is the brother of Elizabeth, who is the sister of Brittney.\newline
    
    Therefore,
    Brittney is the \underline{\textcolor{red}{granddaughter}} of Morgan\newline

    \textbf{SFT-S$+10k$}:\newline
    The ordered structured triples are:\newline
    Brittney is the sister of Elizabeth.\newline
    Elizabeth is the sister of Robert.\newline
    Robert is the son of Kenneth.\newline
    Kenneth is the brother of Harry.\newline
    Harry is the brother of Valerie.\newline
    Valerie is the daughter of Kevin.\newline
    Kevin is the grandfather of Danielle.\newline
    Danielle is the sister of Donald.\newline
    Donald is the brother of Frances.\newline
    Frances is the daughter of Morgan.\newline
    
    Therefore,
    Brittney is the \underline{niece} of Morgan
    
    \\
    \bottomrule
    \end{tabular}
    }
    \caption{Case analysis from the CLUTTR dataset showing the output of the GPT-4o, SFT-S, and SFT-S$+10k$ systems when using our prompt in Appendix~\ref{app:CLUTRR_op}. The left column displays placeholder filler values in the input prompt, as well as the ground truth labels \texttt{[ANSWER]} (which are not sent in the input prompt). For visualization purposes, the final output is \underline{underlined}, and model errors are highlighted in \textcolor{red}{red}.}
    \label{tab:case1}
\end{table*}

\begin{table*}[t]
    \centering
    \resizebox{\textwidth}{!}{
    \begin{tabular}{p{0.5\textwidth}|p{0.5\textwidth}}
    \toprule
    \textbf{Prompt~\ref{app:CLUTRR_op} Placeholder Value} & \textbf{Prediction} \\
    \midrule
    \texttt{[STORY]} \newline
    Nichole was excited because she was meeting her father, Christian, for lunch. Darren and his father Christian went to look at cars. Pennie took her brother Darren with her to eat pasta at a pasta buffet. Sean, Pennie's brother, joined them at the pasta buffet. Sean's mother, Evelyn, was disappointed that he failed his math class. \newline

    \texttt{[QUERY]}\newline
    What is the relationship of Evelyn to Nichole?\newline

    \texttt{[TRIPLES]} \newline
    Evelyn is the mother of Sean.\newline
    Sean is the brother of Pennie.\newline
    Pennie is the sister of Darren.\newline
    Darren is the son of Christian.\newline
    Christian is the father of Nichole.\newline

    \texttt{[ANSWER]} \newline
    Evelyn is the \underline{mother} of Nichole\newline
    &

    \textbf{SFT-S STD-P:} \newline
    Evelyn is the \underline{\textcolor{red}{grandmother}} of Nichole\newline

    \textbf{SFT-S ETA-P:} \newline
    The ordered structured triples are:\newline
    Evelyn is the mother of Sean.\newline
    Sean is the brother of Pennie.\newline
    Pennie is the sister of Darren.\newline
    Darren is the son of Christian.\newline
    Christian is the father of Nichole.\newline
    
    Therefore,
    Evelyn is the \underline{mother} of Nichole
    \\
    \bottomrule
    \end{tabular}
    }

    \caption{Case analysis from the CLUTTR dataset showing the output of the STF-S STD-P and STF-S ETA-P systems when using our prompt in Appendix~\ref{app:CLUTRR_op}. The left column displays placeholder filler values in the input prompt, as well as the ground truth labels \texttt{[ANSWER]} (which are not sent in the input prompt). For visualization purposes, the final output is \underline{underlined}, and model errors are highlighted in \textcolor{red}{red}.}
    \label{tab:case2}
\end{table*}

\begin{table*}[t]
    \centering
    \resizebox{\textwidth}{!}{
    \begin{tabular}{p{0.5\textwidth}|p{0.5\textwidth}}
    \toprule
    \textbf{Prompt~\ref{app:CLUTRR_op} Placeholder Value} & \textbf{Prediction} \\
    \midrule
    \texttt{[STORY]} \newline
    William and his brother James went fishing by the lake.  Steven, the father of James, joined them later to share some stories.  Steve, the son of Steven, brought along some snacks for everyone.  Stanley, Steve's son, was excited to catch his first fish.  James, being the fun uncle, helped Stanley bait his hook.  Patrice, the sister of James, called to check on their fishing adventure.  Darryl, Patrice's son, asked if he could join them next time.  Eric, Darryl's brother, was already planning what to bring for the trip.  Elsie, the grandmother of Eric, loved hearing about their plans and offered to bake some cookies.  Constance, the daughter of Elsie, decided to help her mother with the baking. \newline

    \texttt{[QUERY]} \newline
    What is the relationship of Constance to William?\newline

    \texttt{[TRIPLES]} \newline
    Constance is the daughter of Elsie. \newline
    Elsie is the grandmother of Eric. \newline
    Eric is the brother of Darryl. \newline
    Darryl is the son of Patrice. \newline
    Patrice is the sister of James. \newline
    James is the uncle of Stanley. \newline
    Stanley is the son of Steve. \newline
    Steve is the son of Steven. \newline
    Steven is the father of James. \newline
    James is the brother of William.\newline

    \texttt{[ANSWER]}\newline
    Constance is the \underline{sister} of William
    \newline
    &

    \textbf{GPT4o STD-P:} \newline
    Constance is the \underline{\textcolor{red}{aunt}} of William.\newline

    \textbf{GPT4o ETA-P:} \newline
    The ordered structured triples are:\newline
    1. William is the brother of James.\newline
    2. Steven is the father of James.\newline
    3. Steve is the son of Steven.\newline
    4. Stanley is the son of Steve.\newline
    5. James is the uncle of Stanley.\newline
    6. Patrice is the sister of James.\newline
    7. Darryl is the son of Patrice.\newline
    8. Eric is the brother of Darryl.\newline
    9. Elsie is the grandmother of Eric.\newline
    10. Constance is the daughter of Elsie.\newline

    Analyzing the relationships:\newline
    - William is the brother of James.\newline
    - James is the brother of Patrice.\newline
    - Patrice is the mother of Darryl and Eric.\newline
    - Elsie is the mother of Constance.\newline
    - Elsie is the grandmother of Eric and Darryl.\newline
    - Therefore, Constance is the daughter of Elsie and the sister of Patrice.\newline

    Since Patrice is the sister of James and William, Constance is also the sister of James and William.\newline

    Therefore,
    Constance is the \underline{sister} of William.
    \\
    \bottomrule
    \end{tabular}
    }
    \caption{Case analysis from the CLUTTR dataset showing the output of the GPT4o STD-P and GPT4o ETA-P systems when using our prompt in Appendix~\ref{app:CLUTRR_op}. The left column displays placeholder filler values in the input prompt, as well as the ground truth labels \texttt{[ANSWER]} (which are not sent in the input prompt). For visualization purposes, the final output is \underline{underlined}, and model errors are highlighted in \textcolor{red}{red}.}
    \label{tab:case3}
\end{table*}

\begin{table*}[t]
    \centering
    \resizebox{\textwidth}{!}{
    \begin{tabular}{p{0.5\textwidth}|p{0.5\textwidth}}
    \toprule
    \textbf{Prompt~\ref{app:StepGame_op} Placeholder Value} & \textbf{Prediction} \\
    \midrule
    \texttt{[STORY]} \newline
    In the cozy living room on a rainy Sunday afternoon, Emma sprawled out on the soft carpet with her collection of colorful letter blocks. She carefully placed Block O at the very bottom of her play area. After some contemplation, she decided that Block A should be high above, watching over the rest. She nestled Block Q just above A, creating a hierarchy of letters.

    As she continued, Emma positioned Block S diagonally to the bottom left of Q, making sure it was perfectly aligned. She then placed Block F directly below S, forming a neat vertical line. Finally, she picked up Block M and placed it to the left of F, leaving a small gap between them as if M needed some personal space. Emma admired her work, satisfied with the arrangement of her playful, lettered world. \newline

    \texttt{[QUERY]} \newline
    What is the relation of the agent M to the agent O?\newline

    \texttt{[TRIPLES]} \newline
     M is directly to the left of F. \newline
     F is directly below S. \newline
     S is to the lower-left of Q. \newline
     Q is directly above A. \newline
     A is directly above O.\newline

    \texttt{[ANSWER]} \newline
     M is \underline{directly to the left} of O.
    &

    \textbf{GPT-4o:} \newline
    The ordered structured triples are:\newline
    1. O is at the very bottom of the play area.\newline
    2. A is high above O.\newline
    3. Q is directly above A.\newline
    4. S is to the lower-left of Q.\newline
    5. F is directly below S.\newline
    6. M is directly to the left of F.\newline

    To determine the relation of M to O, we need to trace the spatial relationships:\newline

    - M is directly to the left of F.\newline
    - F is directly below S.\newline
    - S is to the lower-left of Q.\newline
    - Q is directly above A.\newline
    - A is high above O.\newline

    Therefore, M is \underline{\textcolor{red}{to the lower-left}} of O.\newline

    \textbf{SFT-S$+5K$:} \newline
    Ordered structured triples are:\newline
    M is directly to the left of F. \newline
    F is directly below S. \newline
    S is to the lower-left of Q. \newline
    Q is directly above A. \newline
    A is directly above O.\newline
    
    Therefore,
    M is \underline{directly to the left} of O.\newline

    \textbf{SFT-S:} \newline
    Ordered structured triples:\newline
    M is directly to the left of F. \newline
    F is directly below S. \newline
    S is to the lower-left of Q.\newline
    Q is directly above A. \newline
    A is \textcolor{red}{to the upper-right} of O.\newline
    
    Therefore,
    M is \underline{\textcolor{red}{to the lower-left of}} O.\newline
    \\
    \bottomrule
    \end{tabular}
    }
    \caption{Case analysis from the StepGame dataset showing the output of the GPT-4o, SFT-S, and SFT-S$+5k$ systems when using our prompt in Appendix~\ref{app:StepGame_op}. The left column displays placeholder filler values in the input prompt, as well as the ground truth labels \texttt{[ANSWER]} (which are not sent in the input prompt). For visualization purposes, the final output is \underline{underlined}, and model errors are highlighted in \textcolor{red}{red}.}
    \label{tab:case4}
\end{table*}

\begin{table*}[t]
    \centering
    \resizebox{\textwidth}{!}{
    \begin{tabular}{p{0.55\textwidth}|p{0.5\textwidth}}
    \toprule
    \textbf{Prompt~\ref{app:StepGame_op} Placeholder Value} & \textbf{Prediction} \\
    \midrule
    \texttt{[STORY]} \newline
    In a cozy corner of his room, young Alex decided to build an elaborate arrangement with his colorful toy blocks. He carefully placed Block U in the middle of his play mat, imagining it as the central point of his creation. With a gleeful grin, he positioned Block Y to the top-right of Block U, creating a small, vibrant tower. Next, Alex found Block X and placed it snugly at the lower left of Block U, its bright hues contrasting with the other blocks. He then took Block Q and aligned it parallel to Block X, making sure it was directly above, forming a neat, parallel structure. Determined to add more layers, Alex grabbed Block D and put it on the left side, with Block Q neatly set on the right. He stacked Block D above Block P, making sure they stood tall in a vertical line. Directly below Block P, he carefully positioned Block L, extending the vertical plane.Continuing his masterpiece, Alex placed Block B directly below Block L and then added Block O above B, maintaining the vertical alignment meticulously. Satisfied with his towering structure, he added Block H below and to the right of Block O, creating an intriguing diagonal extension. Lastly, with a final flourish, Alex placed Block C at the 9:00 position of Block H, completing his intricate design. He sat back and admired his colorful, geometric world, feeling a sense of accomplishment as the afternoon sun cast playful shadows over his creation.\newline

   \texttt{[QUERY]}\newline
    What is the relation of the agent C to the agent Y?\newline

    \texttt{[TRIPLES]} \newline
    C is directly to the left of H. \newline
    H is to the lower-right of O. \newline
    O is directly above B. \newline
    B is directly below L. \newline
    L is directly below P. \newline
    P is directly below D. \newline
    D is directly to the left of Q. \newline
    Q is directly above X. \newline
    X is to the lower-left of U. \newline
    U is to the lower-left of Y.\newline

    \texttt{[ANSWER]} \newline
    C is \underline{to the lower-left} of Y.\newline

    &

    \textbf{SFT-S STD-P:} \newline
    The agent C is \underline{\textcolor{red}{to the lower-right}} of the agent Y.\newline

    \textbf{SFT-S ETA-P:} \newline
    Ordered structured triples: \newline
    C is directly to the left of H.\newline
    H is to the lower-right of O. \newline
    O is directly above B. \newline
    B is directly below L. \newline
    L is directly below P. \newline
    P is directly below D. \newline
    D is directly to the left of Q. \newline
    Q is directly above X. \newline
    X is to the lower-left of U. \newline
    U is to the lower-left of Y. \newline
    
    Therefore,
    C is \underline{to the lower-left of} Y.
    \\
    \bottomrule
    \end{tabular}
    }
    \caption{Case analysis from the StepGame dataset showing the output of the SFT-S STD-P and SFT-S ETA-P systems when using our prompt in Appendix~\ref{app:StepGame_op}. The left column displays placeholder filler values in the input prompt, as well as the ground truth labels \texttt{[ANSWER]} (which are not sent in the input prompt). For visualization purposes, the final output is \underline{underlined}, and model errors are highlighted in \textcolor{red}{red}.}
    \label{tab:case5}
\end{table*}

\begin{table*}[t]
    \centering
    \resizebox{\textwidth}{!}{
    \begin{tabular}{p{0.5\textwidth}|p{0.5\textwidth}}
    \toprule
    \textbf{Prompt~\ref{app:StepGame_op} Placeholder Value} & \textbf{Prediction} \\
    \midrule
    \texttt{[STORY]} \newline
    In a cozy corner of her room, little Emily spread her colorful building blocks on the soft carpet. She carefully positioned Block L in front of her, a bright blue block that stood proudly at the center of her creation. She then gently placed Block T directly below it, lining them up with precision.With a thoughtful look, she picked up Block D, a vibrant yellow block, and placed it above and to the right of Block L. She smiled as she reached for Block G, placing it just to the left of Block D, ensuring they were perfectly aligned.Finally, she grabbed Block S, a cheerful red block, and positioned it above and slightly to the left of Block G. Emily gazed at her masterpiece with satisfaction, each block perfectly placed in her imaginative world of colors and shapes. \newline

    \texttt{[QUERY]}\newline
    What is the relation of the agent S to the agent T?\newline

    \texttt{[TRIPLES]} \newline
    S is to the upper-left of G. \newline
    G is directly to the left of D. \newline
    D is to the upper-right of L. \newline
    L is directly above T. \newline

    \texttt{[ANSWER]}\newline
    S is \underline{to the upper-left} of T.\newline

    &

    \textbf{GPT4o STD-P:} \newline
    S is \underline{\textcolor{red}{to the upper-right}} of T.\newline

    \textbf{GPT4o ETA-P:} \newline
    The ordered structured triples are: \newline
    1. T is directly below L. \newline
    2. D is to the upper-right of L. \newline
    3. G is directly to the left of D. \newline
    4. S is to the upper-left of G.  \newline

    To determine the relation of S to T, we need to trace the spatial relationships:  \newline
    - S is to the upper-left of G. - G is directly to the left of D. \newline
    - D is to the upper-right of L. \newline
    - T is directly below L.  \newline

    Since D is to the upper-right of L, and G is directly to the left of D, G is to the upper-left of L. S is to the upper-left of G, which places S further to the upper-left of L. Given that T is directly below L, S is positioned above and to the left of T. \newline
    
    Therefore,
    S is \underline{to the upper-left} of T.
    \\
    \bottomrule
    \end{tabular}
    }
    \caption{Case analysis from the StepGame dataset showing the output of the GPT4o STD-P and GPT4o ETA-P systems when using our prompt in Appendix~\ref{app:StepGame_op}. The left column displays placeholder filler values in the input prompt, as well as the ground truth labels \texttt{[ANSWER]} (which are not sent in the input prompt). For visualization purposes, the final output is \underline{underlined}, and model errors are highlighted in \textcolor{red}{red}.}
    \label{tab:case6}
\end{table*}

\end{document}